\pdfoutput=1

\documentclass[conference]{IEEEtran}
\IEEEoverridecommandlockouts
\usepackage{cite}
\usepackage{amsmath,amssymb,amsfonts}
\usepackage{algorithmic}
\usepackage{graphicx}
\usepackage{textcomp}
\usepackage{xcolor}
\def\BibTeX{{\rm B\kern-.05em{\sc i\kern-.025em b}\kern-.08em
    T\kern-.1667em\lower.7ex\hbox{E}\kern-.125emX}}

\usepackage{listings}
\usepackage{booktabs}
\usepackage{multirow}
\usepackage{subfigure}
\usepackage{caption}

\usepackage{arydshln}
\usepackage{multirow}
\usepackage{multicol}
\usepackage{bm}
\usepackage{url}
\usepackage{underscore}
\usepackage{color}
\usepackage{colortbl}
\usepackage{makecell}
\usepackage{enumitem}
\usepackage{soul}
\usepackage{arydshln}
\usepackage{amssymb}

\newcommand{\cz}[1]{{\color{black}#1}}
\newcommand{\hyf}[1]{{\color{black}#1}}

\newcommand{\gyx}[1]{{\color{black}#1}}

\newcommand{\wen}[1]{{\color{black}#1}}

\newcommand{\rmnum}[1]{\romannumeral #1}
\newcommand{\Rmnum}[1]{\expandafter\@slowromancap\romannumeral #1@}

\begin{document}
\title{
Tele-Knowledge Pre-training for Fault Analysis}

\author{\IEEEauthorblockN{\normalsize Zhuo Chen$^\dagger$,Wen Zhang$^\dagger$, Yufeng Huang, Mingyang Chen,Yuxia Geng, Hongtao Yu, Zhen Bi, Yichi Zhang, Zhen Yao  }
\IEEEauthorblockA{
{\normalsize \textit{Zhejiang University}, 
Hangzhou, China} \\
{\small \{zhuo.chen, zhang.wen, huangyufeng, mingyangchen, gengyx, yuhongtaoaaa, bizhen_zju, 22221092, 22151303\}@zju.edu.cn} 
}
\and
\IEEEauthorblockN{\normalsize Wenting Song, Xinliang Wu, Yi Yang, Mingyi Chen, Zhaoyang Lian, Yingying Li, Lei Cheng}
\IEEEauthorblockA{
{\normalsize \textit{NAIE PDU, Huawei Technologies Co., Ltd.}, 
Xi'an, China} \\
{\small\{songwenting, wuxinliang1, yangyi193, chenmingyi2, lianzhaoyang, liyingying66, chenglei\}@huawei.com }
}
\and
\IEEEauthorblockN{\normalsize Huajun Chen$^\ddagger$}
\IEEEauthorblockA{
{\normalsize \textit{Zhejiang University}} \\
{\small huajunsir@zju.edu.cn}
}
}

\maketitle

\def\thefootnote{$\dagger$}\footnotetext{Equal Contribution.}
\def\thefootnote{$\ddagger$}\footnotetext{Corresponding Author.}

\begin{abstract}
In this work, we share our experience on tele-knowledge pre-training for fault analysis,
\cz{a crucial task in telecommunication applications that requires a wide range of knowledge normally found in both machine log data and product documents. }
\cz{
To organize this knowledge from experts uniformly, we propose 
\wen{to create} a Tele-KG (tele-knowledge graph). 
Using this valuable data, we further propose a 
tele-domain 
\wen{language}
pre-training model \textbf{TeleBERT} and its knowledge-enhanced version, 
\wen{a tele-knowledge re-training model}
\textbf{KTeleBERT}. 
\wen{which} 
includes effective prompt hints, adaptive numerical data encoding, and two knowledge injection paradigms.
Concretely, 
\wen{our proposal includes}
two stages: first, pre-training TeleBERT on 20 million tele-related corpora, and then re-training it on 1 million causal and machine-related corpora to obtain KTeleBERT.}
\cz{Our evaluation on multiple tasks related to fault analysis in tele-applications, including root-cause analysis, event association prediction, and fault chain tracing, 
shows that pre-training a language model with tele-domain data is beneficial for downstream tasks. 
Moreover, 
 the KTeleBERT re-training further improves the performance of task models, highlighting the effectiveness of incorporating diverse tele-knowledge into the model.
}

\end{abstract}

\begin{IEEEkeywords}
telecommunication, model pre-training, knowledge graph, numeric encoding, fault analysis 
\end{IEEEkeywords}

\section{Introduction} 
Faults in telecommunication networks (tele-network) can have a major impact on the availability and effectiveness of the global network, resulting in significant maintenance costs for operating companies. Thus, quick elimination of the faults and preventing the causes of fault generation are crucial for the special interest of operating companies. 
Fault analysis is a complex task composed of multiple sub-tasks, requiring a wealth of tele-knowledge such as the network architecture and the dependence among tele-products. Historically, this knowledge was stored in the minds of experts. While in now-days, massive product data and expert experience in tele-field are accumulated in various forms.
For example, as the valuable first-hand data, the \textbf{machine (log) data} (e.g., abnormal event like the alarm or normal indicator like the KPI score) is raised continuously in both real tele-scenario and laboratory environments. 
  \begin{figure}[!htbp]
  \centering
  \includegraphics[width = 0.70\linewidth]{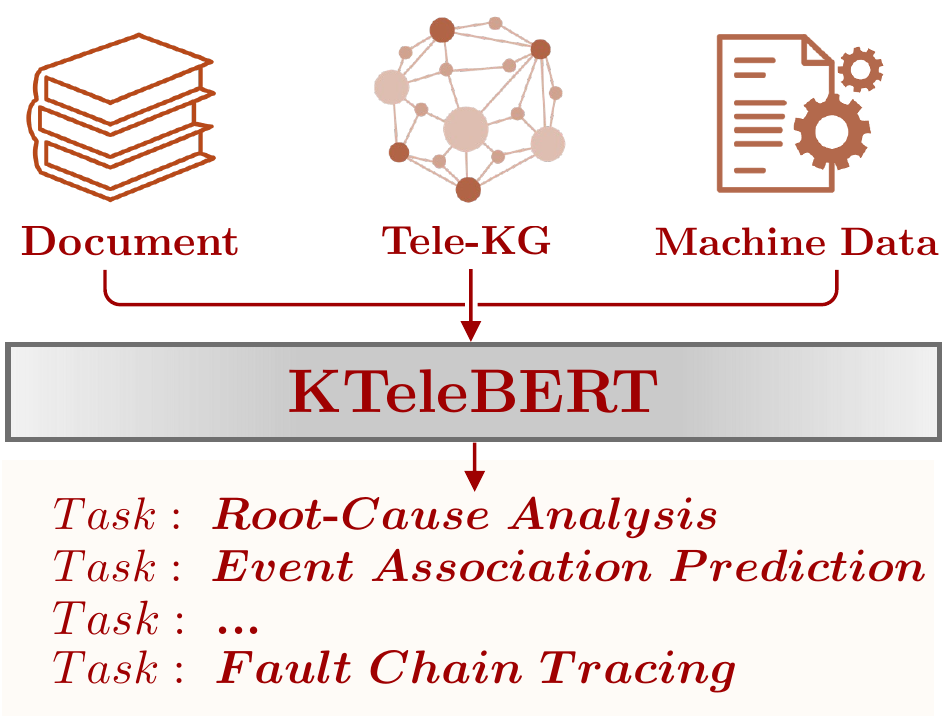}
  \vspace{-2pt}
  \caption{Workflow for our KTeleBERT.}
  \label{fig:intro}
  \vspace{-14pt}
\end{figure}
\cz{Additionally, the \textbf{product documents} are created for tele-products in the network, containing detailed information such as the product profile, event description, fault case, and solutions to particular issues, primarily in natural language.}

Nevertheless, some knowledge such as the the types of faults and their hierarchy, is still not uniformly recorded. Considering the diversity of such knowledge, knowledge graph (KG) is a common choice to represent them, which represents facts as triples, such as \emph{(China, capitalIs, Beijing)}.
\wen{In} recent years, KGs have been widely adopted in industry \cite{DBLP:conf/icde/ZhangWYWZC21,DBLP:conf/jist/ZhangDCWCXLC21,DBLP:conf/mm/ZhuZZYCZC21} due to their flexibility and convenience  to easily combine data from multiple sources. 
To uniformly represent the recorded tele-knowledge, we built a tele-product knowledge graph (a.k.a. \textbf{Tele-KG}). 
For example, given the triple \emph{([Alm] ALM-$100072$ The NF destination service is unreachable, trigger, [KPI] $1929480378$ The number of initial registration requests increases abnormally)}, 
 it represents that the Network Function (NF) destination service being unreachable (alarm $100072$) always results in the number of initial registration requests increasing (KPI abnormal event $1929480378$). 
 We note that the majority of knowledge in Tele-KG is derived from experts and engineers, providing an integrated view of tele-knowledge and accumulated experience.

\cz{Although the Tele-KG can be used as a knowledge base to retrieve knowledge using SPARQL queries \cite{DBLP:conf/icdt/Schmidt0L10} for simple fault analysis support, this 
\wen{solution}
is still inflexible and 
have limitations in generalization capabilities to those indirectly associated tasks.}
Another way to utilize Tele-KG is through knowledge graph embedding (KGE) methods \cite{DBLP:conf/nips/BordesUGWY13,DBLP:conf/aaai/WangZFC14,DBLP:conf/icml/TrouillonWRGB16,DBLP:conf/iclr/SunDNT19}, which aims to learn embeddings of entities and relations in a continuous vector space and then assist the knowledge inference like the task of link prediction or triple classification in a KG.
\cz{However, those technologies always suffer from the knowledge inconsistency, i.e., the same entity or noun in the real world may have different surfaces like the ``\emph{Alm}'' v.s. ``\emph{Alarm}''.
Besides, the textual knowledge and semantic information in entity surfaces are always abandoned during training, limiting models' intra-domain scalability and cross-domain portability.}

\cz{The textual product documents are valuable resources in tele-domain. Instead of simply using them as handbooks, one approach is to pre-train a domain-specific language model (LM).}
LM pre-training \cite{DBLP:conf/naacl/DevlinCLT19,DBLP:journals/corr/abs-1904-09223,radford2018improving,DBLP:journals/corr/abs-1907-11692} is a good recipe for learning implicit semantic knowledge 
with self-supervised text reconstruction as the training objective in a vast amount of language data.
However, 
their challenges lie in exploiting the structured knowledge for explicit intellectual reasoning.
Additionally, our machine data is semi-structured and multi-directional: with a vertical direction of the time and a horizontal direction of multiple indicators extending the machine data at a single moment, as shown in Fig. \ref{fig:TeleKG}(a). This differs from the typically log-based anomaly detection methods \cite{DBLP:conf/ijcai/GuoTYLH17,DBLP:conf/cikm/SongS0DX0T19,DBLP:conf/kdd/GuoCTZLH21} which target at the unidirectional and serial log data.

In this work, we propose to pre-train all data that contains tele-knowledge, including machine data, Tele-Corpus from the product documents, and triples from the Tele-KG.
\cz{We expect that this pre-trained model can aid in downstream fault analysis tasks in a convenient and effective manner, and boost their performance, especially for tasks with limited data (also known as low-resource tasks)}.

\cz{To achieve this, we first address the issue from multi-source and multi-modal data (e.g., multi-directional machine data, textual documents, and semi-structured KG), which can distract the model from efficient learning. To remedy this, we refer to the \textbf{prompt engineering} techniques \cite{DBLP:journals/corr/abs-2107-13586,DBLP:conf/acl/GaoFC20,DBLP:conf/www/ChenZXDYTHSC22} for modality unification and provide relevant \textbf{template hints} to the model for modalities unification.}

\cz{Secondly, we address the challenge of handling numerical data, which is an essential component of data in tele-domain and frequently appears in machine data (e.g., KPI scores).} 
This data format is similar to the tabular data, sharing the characteristic of: 
\textbf{\textit{(\rmnum{1})}} The text part is short;  \textbf{\textit{(\rmnum{2})}} The Numerical values always have different meanings and ranges under different circumstances;
\textbf{\textit{(\rmnum{3})}} Data stretches from both vertically and horizontally which is hierarchical.
However, existing table pre-training methods mainly study the hierarchical structure of tabular data \cite{DBLP:conf/ijcai/0001CHZZZLHZ22,DBLP:conf/acl/YinNYR20,DBLP:conf/iclr/LiuCGZLCL22,DBLP:conf/naacl/IidaTMI21,DBLP:conf/coling/GongSFQBLL20,DBLP:conf/acl/YangGUHGP22} where the numerical information is rarely studied in depth.
Furthermore, those methods that target at learning numerical features \cite{DBLP:conf/ijcai/GuoTYLH17,DBLP:conf/cikm/SongS0DX0T19,DBLP:conf/kdd/GuoCTZLH21} focus on learning field embedding  for each numerical field. 
\cz{They tend to consider the task with limited fields (e.g., the user attributes like height and weight) but fail when migrated to our tele-scenario where the field number (e.g., KPI name) is numerous ($\geq$$1000$) and new fields are often generated during the development of enterprise.}
Thus, we propose an adaptive numeric  encoder (ANEnc) in tele-domain for type-aware numeric encoding.

\cz{Thirdly, we are aware of different training target among the tele-corpus, machine data and the knowledge triples.} Thus
we adopt a multi-stage training mode for multi-level knowledge acquisition:
\textit{(i)} \textbf{TeleBERT}: in stage one we follow ELECTRA \cite{DBLP:conf/iclr/ClarkLLM20} pre-training paradigm and data augmentation method SimCSE \cite{DBLP:conf/emnlp/GaoYC21} for large-scale (about 20 million) textual tele-corpus pre-training;  
\textit{(ii)} \textbf{KTeleBERT}:  In stage two, we extract those causal sentences which contain relevant causal keywords to re-train TeleBERT together with the numeric-related machine data, where a knowledge embedding training objective and multi-task learning method are introduced for explicit knowledge integration. 

With our pre-trained model, 
we apply the model-generated service vectors to enhance three tasks of fault analysis: root-cause analysis (RCA), event association prediction (EAP), and fault chain tracing (FCT). 
The experimental results show that our TeleBERT and KTeleBERT successfully improve the performance of these three tasks.

In summary, \hyf{the} contributions of this work are as follows:
\begin{itemize}
	\item \wen{We emphasize the importance of encoding knowledge uniformly in tele-domain application, and share our encoding experience in real-world scenarios.}
	\item We propose a tele-domain pre-training model TeleBERT and its knowledge-enhanced version KTeleBERT to fuse and encode diverse tele-knowledge in different forms.
	\item We prove that our proposed models could serve multiple fault analysis task models and  boost their performance.
\end{itemize}

\section{Background} 

\subsection{Corpus in Telecommunication}
 \begin{figure*}[htbp]
 \centering
 \includegraphics[width = 0.97\linewidth]{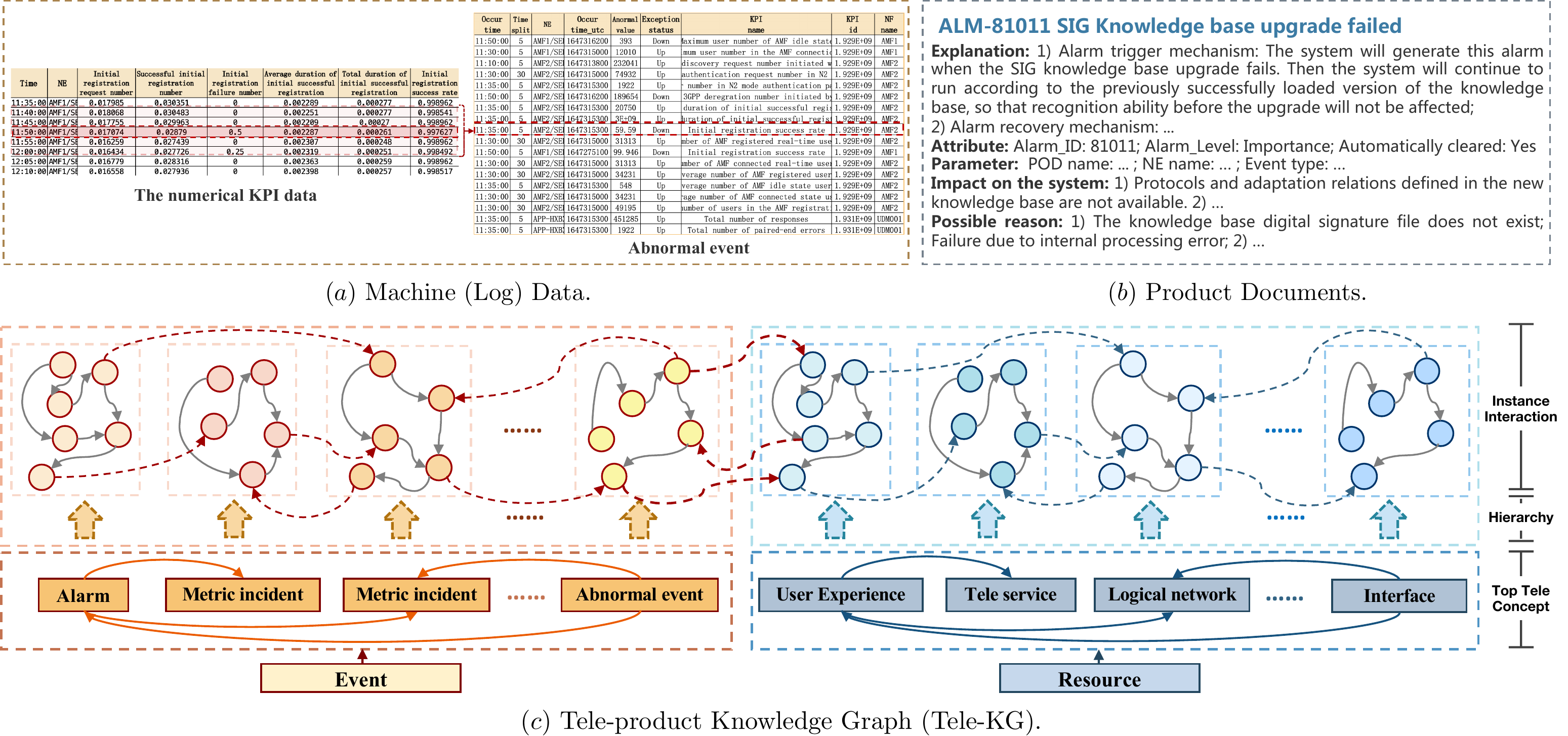}
 \caption{Corpus overview where all the Chinese corpus are translated into English to improve the comprehensibility.}
 \label{fig:TeleKG}
 \vspace{-10pt}
\end{figure*}
\subsubsection{Machine Log Data}
\cz{The machine (log) data, such as abnormal events or normal indicator logs, is continuously generated in both real-world tele-environments and simulation scenes. 
Typically, as shown in \ref{fig:TeleKG}(a),
these abnormal events like the service interruption, have varying levels of importance and are always accompanied by anomalies in  relevant network elements (NEs). 
The normal indicators like the numerical KPI data, on the other hand, are cyclical and persistent in nature and make up the majority of automatically generated machine data.
}
\cz{Most abnormal events can self-recover after existing a period of time, (e.g., network congestion), and there may be correlation or causal relationships across abnormal events or indicators}, e.g, the alarm ``\emph{(NF destination service is unreachable)}'', always lead to abnormal KPI score ``\emph{(the number of initial registration requests increases abnormally)}''.

\subsubsection{Product Document}
Those domain engineers or experts are constantly recording and updating the \textbf{product documentation}. 
Particularly, each scenario may contain one or more product documents, which are maintained by different departments and may include nearly all relevant information in the field, such as the fault cases, solutions for already occurred or potential cases, and the event descriptions shown in \ref{fig:TeleKG}(b).

\subsubsection{Tele-product Knowledge Graph (Tele-KG)}
We construct the Tele-KG to integrate massive information about events and  resources on our platform.
Our goal is intuitive: hoping that such a fine-grained Tele-KG could refine and purify the knowledge of tele-domain, \cz{as a semi-structured knowledge graph is more flexible and has higher knowledge density than traditional structured databases or unstructured product documents.}
Specifically, we define a hierarchical 
\wen{tele-schema}
as the guidance for KG construction, as shown in Fig. \ref{fig:TeleKG}(c), where the top-down modeling method is adopted for schema design. The concept classes across different levels are inherited via ``\textit{subclassOf}'', and those classes within the same levels are connected via common relations like ``\textit{provide}''.

We note that top superclasses ``\textit{Event}'' and ``\textit{Resource}'' are defined as the root in tele-domain,
with other top tele-concept as the subdivisions. 
\cz{The instantiation of the tele-schema at instance level contains interactions among different instances and forms the majority of the Tele-KG, including those triple cases mentioned before.}

\subsection{Task of Fault Analysis}

\subsubsection{Root-Cause Analysis}
\cz{In modern telecommunication systems, the identification of the root causes of abnormal events is essential for reducing financial losses and maintaining system stability. }
\cz{However, traditional methods of root-cause analysis rely heavily on manual work by experts, using summarized documents meanwhile incurring significant financial and human resources. As the size and complexity of these systems continue to grow, manual analysis becomes increasingly difficult. Therefore, developing an automated method for root-cause analysis is a pressing need in tele-domain.}

\subsubsection{Event Association Prediction}
\cz{One approach for finding the root cause of a fault event is to utilize prior trigger relationships between different fault events. These relationships can reveal patterns of fault causation, such as a triple (\textit{Alarm A, triggers, Alarm B}) indicating that the \textit{Alarm B} is caused by the \textit{Alarm A}. By traversing these trigger relations, the root cause of a current fault event can be determined. 
However, traditionally, these trigger relationships have been identified by tele-experts through manual analysis of a large number of fault cases, which is time-consuming and 
\wen{is}
limited by personal bias.
This is also difficulty in updating or adapting to new network changes.
}
\wen{Thus proposing effective methods  
for automatically predicting the trigger relationship in candidate event pairs is important.}

\subsubsection{Fault Chain Tracing}
\cz{
Network equipment failure is a common phenomenon in tele-domain due to high operating pressure of the network. In these failure scenarios, alarms are often raised, which can have a cascading effect and cause damage to the entire system. Tracing the source of these failures is crucial for maintaining the stability of the tele-network. Traditionally, this task is accomplished by experts with their experience, sharing the limitations with the above two tasks. Therefore, developing an automated method for fault chain tracing is quite valuable and necessary.}

\section{Pre-training on Tele-communication corpora} 

In this section we introduce our TeleBERT, a tele-domain specific PLM  pre-trained on large-scale textual Tele-Corpus.

\subsection{Telecommunication Corpus Integration} \label{sec:telebert}
\wen{The}
large-scale textual telecommunication corpora 
\wen{consists}
of sentences from various sources, including product documents and entity surfaces within the Tele-KG. 
To expand the dataset and increase the diversity of the training data, we apply two data augmentation techniques from the NLP community:
\textbf{\textit{(\rmnum{1})}} Explicit data augmentation:
\cz{we splice together a range of adjacent sentences from the same document to expand the dataset and create a final pre-training corpus of 20 million sentences (a.k.a. \textbf{Tele-Corpus}).}
\textbf{\textit{(\rmnum{2})}} Implicit data augmentation: 
\cz{following SimCSE \cite{DBLP:conf/emnlp/GaoYC21}, we introduce noise into the dataset through a dropout strategy to enhance the robustness of our model.}

\subsection{TeleBERT} 
\cz{The pre-training process follows the vanilla mask language model (MLM) \cite{DBLP:conf/naacl/DevlinCLT19} where each sentence is fed to the model with a special token {\tt [CLS]} prepended and {\tt [SEP]} appended.}
Taking the Chinese pre-trained language model (PLM) MacBERT \cite{DBLP:journals/taslp/CuiCLQY21} as the backbone, we adopt the whole word masking (WWM) strategy during TeleBERT pre-training.
\cz{It segments the text into the ``whole words" using a tele-domain vocabulary of approximately $372k$ Chinese or English elements that are mostly proper nouns (e.g., ``\textit{QoS}'' which refs to ``\textit{Quality of Service}'') or phrases (e.g., ``\textit{network congestion points}'' and ``\textit{dedicated control channel}'').}

Moreover, the simple contrastive learning on sentence embeddings (SimCSE) \cite{DBLP:conf/emnlp/GaoYC21} is employed to alleviate the collapse of representation learning on large models, i.e., most sentences are represented by similar embeddings.
The ELECTRA \cite{DBLP:conf/iclr/ClarkLLM20} pre-training paradigm is used for increasing pre-training difficulty, where a MLM generator is equipped for mask reconstruction and make{s} the TeleBERT a discriminator with a self-supervised objective of replaced token detection (RTD) applied.  
Note that we define the TeleBERT pre-training progress as the stage one, aiming to let the PLM understand the general semantic knowledge in tele-domain.

\section{Re-training on Causal and Machine Corpora} \label{sec:retrain}

\cz{In this section, we introduce the stage two where TeleBERT is re-trained to create the KTeleBERT model. Specifically, in Sec. \ref{sec:Unify}, we describe the rules for causal sentences extraction and how to efficiently unify those multi-modal data for model re-training. Then, we elaborate our proposed numerical data encoding module (Sec. \ref{sec:num}) and different strategies for mask reconstruction (Sec. \ref{sec:mask}).
Finally, we introduce our method for explicit expert knowledge injection (Sec. \ref{sec:kge}) and introduce the training policy for task integration (Sec. \ref{sec:trainstrategy}).}

\subsection{Unifying Modalities and Patterns} \label{sec:Unify}
\subsubsection{Causal sentences extraction}
\cz{
After removing unique identifiers like `` \textit{[KPI] 1929480378}'', 
\textbf{\textit{(\rmnum{1})}} we manually select words and phrases with causal meanings, such as ``\textit{affect}'' and ``\textit{lead to}'', as the causal keywords. We also heuristically customize extraction rules (e.g., minimum length) to obtain approximately $200k$ sentences from the Tele-Corpus that both contain these causal keywords and satisfy the rule constraints. 
Then, \textbf{\textit{(\rmnum{2})}} relational triples and attribute triples (which contain crucial attribute after evaluated) from the Tele-KG are serialized by concatenating the surfaces of entities/attributes and relations  for sequential format unification. This process serves as \textbf{a manner for implicit knowledge injection} \cite{DBLP:journals/corr/abs-2207-12888,DBLP:conf/acl/ZhangHLJSL19}}.

\subsubsection{Prompt template construction}
\begin{figure}[!htbp]
  \centering
  \vspace{-5pt}
  \includegraphics[width = 0.94\linewidth]{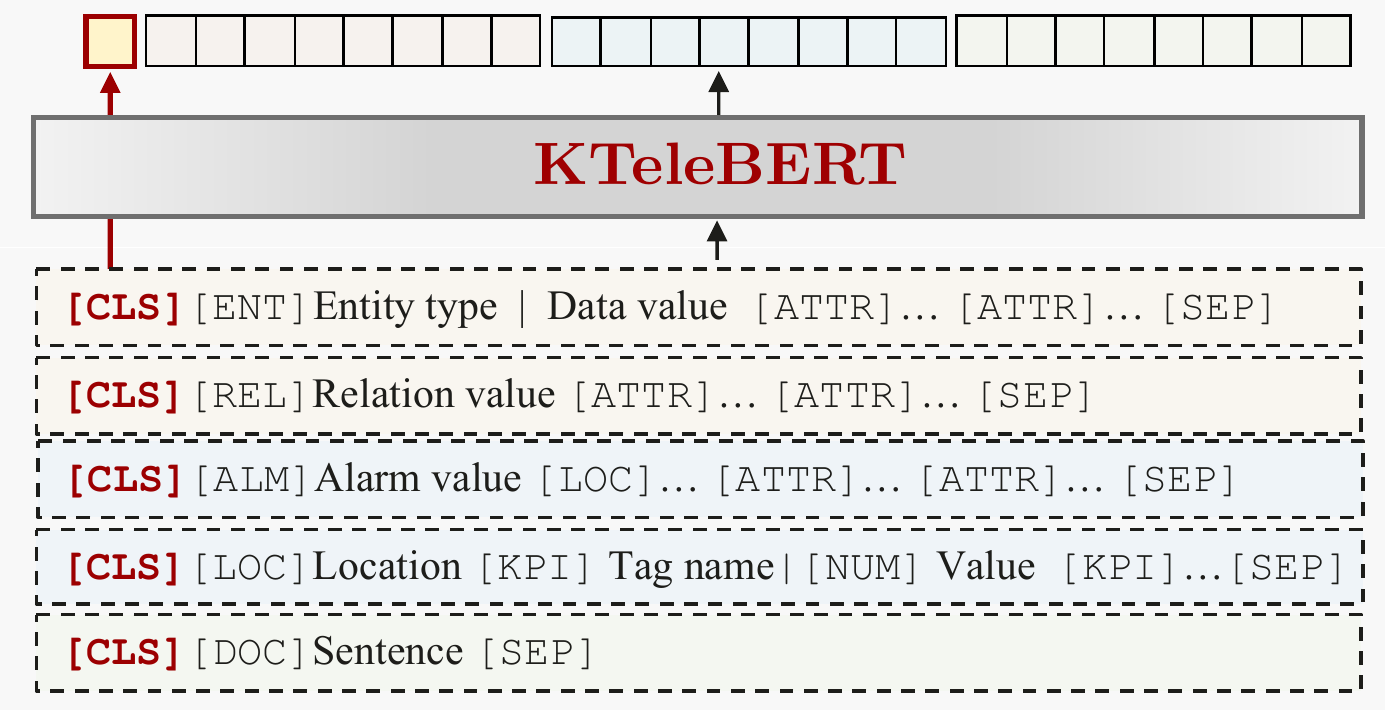}
  \caption{Prompt template for KTeleBERT. Those corpora are  wrapped with our prompt templates to unify the input modality. 
  The full names are as follows: Alarm ({\tt [ALM]}); Relation ({\tt [REL]}); Entity ({\tt [ENT]}); Location ({\tt [LOC]}); Document ({\tt [DOC]}); Attribute ({\tt [ATTR]}); Numeric ({\tt [NUM]}).
  }
  \label{fig:KTeleBERT}
  \vspace{-6pt}
\end{figure}

\cz{To handle the disordered nature of the structured machine (log) data and the attribute-equipped entity in knowledge triples, we introduce \textbf{special prompts} (tokens) to represent the category of the immediately following content, which has been proved beneficial in knowledge utilizing and cross-modal learning \cite{DBLP:journals/corr/abs-2107-13586,DBLP:conf/acl/GaoFC20,DBLP:conf/www/ChenZXDYTHSC22,DBLP:journals/corr/abs-1909-03193}. For example, {\tt [ATTR]} indicates that the following context is an attribute with its value. We wrap the input with our prompt template to unify the data modality which alleviates the disorder issue brought by the structured machine or attribute data. To further distinguish different attribute types, we use the symbol ``{\tt $|$}'' to split the type names and their values. The pre-defined prompt templates are illustrated in Fig. \ref{fig:KTeleBERT}.}

\subsubsection{Tele special token construction}
\cz{In order to further improve model's understanding of domain-specific words and phrases, we adopt the Byte Pair Encoding (BPE) \cite{DBLP:conf/acl/SennrichHB16a} algorithm to merge characters and learn a collection of tele-specific tokens. This is done by iteratively counting all symbol pairs and selecting the most frequent ones, with a pre-defined symbol vocabulary size as the constraint. 
We find that these candidate tokens, which are mostly significant \textbf{abbreviations of domain-specific phrases or nouns}, satisfy the following two constraints: \textbf{\textit{(\rmnum{1})}} The length of the character sequence is between $2\sim4$, and \textbf{\textit{(\rmnum{2})}} the token appeared frequently (e.g., more than $8000$ times) in the Tele-Corpus and is not included in original MacBERT/BERT vocabulary. Examples of filtered tokens include ``{\tt RAN}'', ``{\tt MML}'', ``{\tt PGW}'', ``{\tt MME}'', ``{\tt SGW}'', ``{\tt NF}''. These tele-specific tokens, along with the prompt tokens, are all inserted into the vocabulary of KTeleBERT as special tokens with newly learnable token embeddings added. 
}

\subsection{Numerical Data Encoding} \label{sec:num}
We claim that the major information contained in machine data comes from its numerical value paired with the tag (type) name, which indicates the meaning of the corresponding numeric data. Different numerical values often have implied associations, as reflected in their synergistic fluctuations in value. 
\cz{For example, an abnormal increase in the number of ``\textit{PDU Session Establishment Reject}'' messages on interface ``\textit{N11}'' 
may lead to a sudden decrease in the successful rate of ``\textit{5G SA Session Establishment}''.
The correlations among various numerical information are a valuable supplement to expert experience in the tele-domain, as the machine data is constantly generated. Moreover, some numerical data can also be found in attribute triples from the Tele-KG.} 

\begin{figure}[htbp]
  \centering
  \vspace{-2pt}
  \includegraphics[width = 0.5\linewidth]{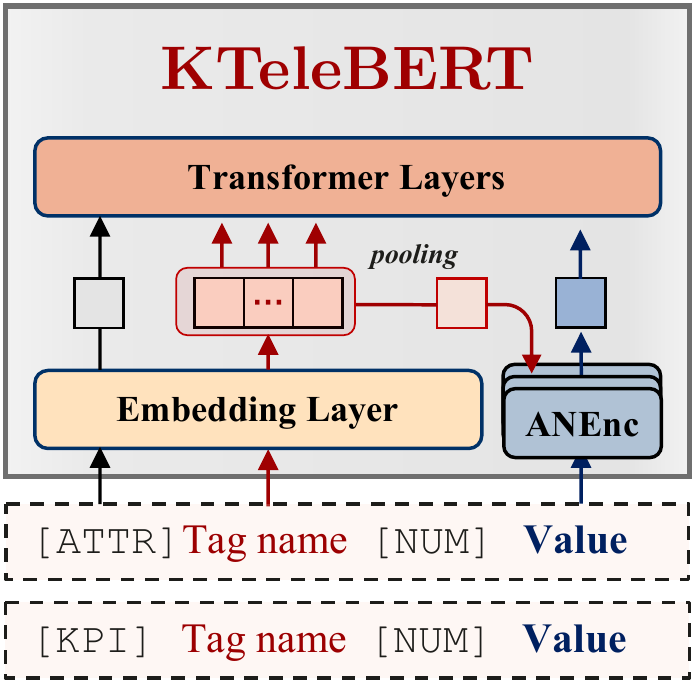}
  \caption{Numerical value encoding in KTeleBERT.}
  \label{fig:num}
  \vspace{-5pt}
\end{figure}

As elaborated in Sec. \ref{sec:Unify}, we unify the modalities and patterns of the input to alleviate the problems caused by inconsistent input formats.
We noticed that existing numerical learning methods \cite{DBLP:conf/ijcai/GuoTYLH17,DBLP:conf/cikm/SongS0DX0T19,DBLP:conf/kdd/GuoCTZLH21} 
always fail when migrated to our tele-scenario where the field number (e.g., KPI name) is numerous and new names are often generated.
To address this issue, we design an \textbf{adaptive numeric encoder} (ANEnc) module to encode the fine-grained numerical data  meanwhile adapt to numerous fields. 
\cz{The ANEnc module is an integral part of our KTeleBERT}  as shown in  Fig. \ref{fig:num}.

\begin{table}[ht]
	\centering  
	\caption{Part of the notations and symbols used in Sec. \ref{sec:retrain}.}
	\begin{tabular}{ll}
	\toprule
		\makebox[0.8cm][c]{\textbf{Symbols}}  & \makebox[6.5cm][c]{\textbf{Description}}  \\
	\midrule
		$t$ & $d$-dimensional tag name embedding  \\
		$v^{tag}$  & Numerical value $v$ with $tag$ as the tag name \\
		$N$  & Number of field aware meta embedding in each layer \\ 
		$\hat{h}$ & Output for attention-based numeric projection (ANP)\\
		$h^l$ & Output for $l$-th adaptive numeric encoder (ANEnc) layer\\
		$x$ & Input numerical embedding for each ANEnc layer\\
		$\bm{E}$  & $N \times (d/N)$ Matrix collection for meta embeddings \\
		$e^{(i)}$  & The $i$-th meta embedding in $\bm{E}$ \\
		$\bm{W}_q$ &  $d \times (d/N)$ matrix for Query conversion \\
		$\bm{W}_v^{(i)}$  & $d \times d$ matrix for numeric transformation \\
		$q$ & query embedding which equals to $t\bm{W}_q$ \\
	\bottomrule
	\end{tabular} 
	\label{tab:param}
	\vspace{-7pt}
\end{table}

 \begin{figure*}[htbp]
  \centering
  \vspace{-2pt}
  \includegraphics[width = 0.72\linewidth]{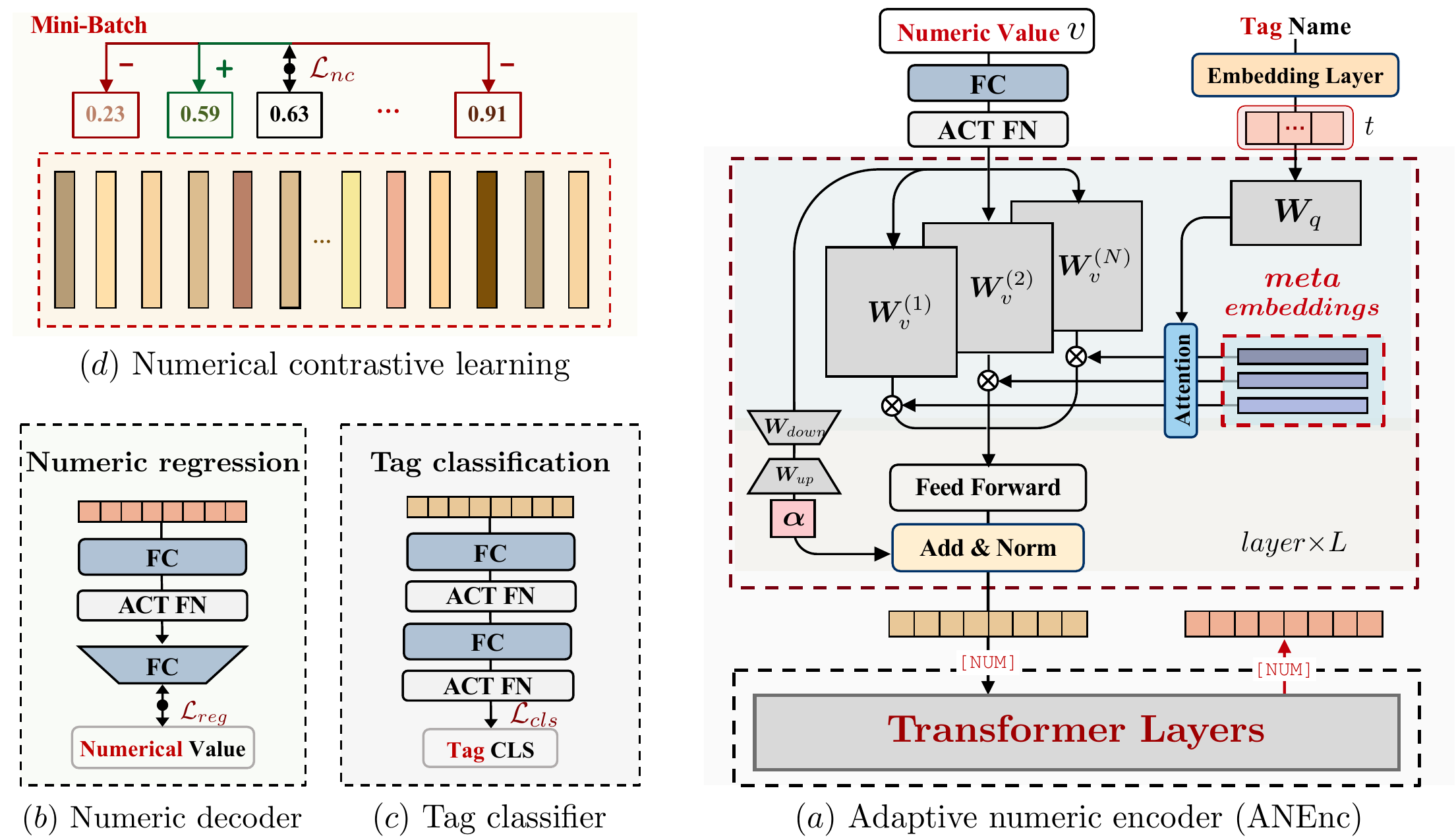}
  \caption{Framework for adaptive numeric encoding process.}
  \label{fig:ANE}
  \vspace{-10pt}
\end{figure*}
The whole process for our numeric encoding model, as shown in Fig. \ref{fig:ANE}(a), are composed of $L$ stacked ANEnc layers together with a numeric decoder (NDec) module. 
Each of the ANEnc layer contains two types of sub-layers: attention-based numeric projection (ANP) and a fully connected feed-forward network (FFN). 
Specifically, we construct $N$ learnable field aware meta embeddings $\bm{E} \in \mathbb{R}^{N \times (d/N)}$. 
Each of the meta embedding $e^{(i)}$ is paired with a \textit{Value} conversion function parameterized by $\bm{W}_v^{(i)}  \in \mathbb{R}^{d \times d}$.
Note that the meta embedding $e^{(i)}$ denotes one decoupled aspect of the domain knowledge and the conversion matrix $\bm{W}_v^{(i)}$  represents the numerical embedding transformation manner in this meta domain.
Moreover, we define the \textit{Query} projection as $\bm{W}_q  \in \mathbb{R}^{d \times (d/N)}$, which converts the $d$-dimensional tag name embedding $t$ into a ($d/N$)-dimension query embedding $q$. Then the attention score $s_{attn}^{(i)}$ for each  meta domain $i$ is calculated by attention function using $(q, e^{(i)})$, 
and the output of each projection matrix $\bm{W}_v^{(i)}$ are summed by attention-based fractional weighting  to get the domain-adaptive embedding $h$.
Note that the embedding $t$ remains unchanged across all ANEnc layer, which is tag name's  pooling output embedding from the former embedding layer.
We following previous works \cite{DBLP:conf/ijcai/GuoTYLH17,DBLP:conf/cikm/SongS0DX0T19,DBLP:conf/kdd/GuoCTZLH21}  to put a single fully connected network with $\bm{W}_{fc} \in \mathbb{R}^{1 \times d}$ prepended for numerical value mapping. Note that all numerical values across the same tag name should be normalized via Min-max normalization, as shown in Fig. \ref{fig:TeleKG}(a), to smooth the learning process.
The above process could be represented as:
\begin{equation} 
\hat{h} = 
softmax(\frac
{t W_q E^T}
{\sqrt{d/N}}
)
 V\, ,
\end{equation}
where 
\begin{equation}
    V = (x \bm{W}_v^{(i)} ~ \cdots ~  x \bm{W}_v^{(N)})\, .
\end{equation}
Note that in the $l$-th ANEnc layer ($l \leq L$) 
\begin{equation} 
x =
\begin{cases}
\text{{ACT}$\_${FN}} ( v \bm{W}_{fc}) & \l=1\\
h^{(l-1)} & otherwise \,
\end{cases}
\end{equation}
where 
\begin{equation}
  \vspace{-1pt}
h = 
 Norm(FFN(\hat{h}) + \alpha \cdot x W_{down} W_{up}) \, ,
   \vspace{-1pt}
\end{equation}
which is the output hidden state of the following FFN sub-layer with trainable low-rank matrices injected to approximate the weight update, following LoRA \cite{DBLP:conf/iclr/HuSWALWWC22}. 
Concretely, $\bm{W}_{down} \in \mathbb{R}^{d \times r}$, and $\bm{W}_{up} \in \mathbb{R}^{r \times d}$ are turnable parameters where $r$ $\leq$ $d$. $\alpha$ $\geq$ $1$ is a  turnable scalar hyperparameter. 
Then, the output embedding $h^{L}$ (represented by $h$ in following context) is fed into the following stacked transformers layer together with those normal token embeddings. 

\subsubsection{Numeric regression}
In order to make the ANEnc compatible with the typical self-supervised pre-training mode, we introduce a numeric decoder (NDec) module to form an autoencoder-like framework.
Concretely, 
\cz{the NDec module takes the output embedding of the final transformer layers as input, allowing for the incorporation of semantic interactive information across multiple transformer layers.} 
Assuming NDec's $1$-dimensional  embedding output as $v_p$. 
As shown in Fig. \ref{fig:num}(b), we present the numeric regression loss
$\mathcal{L}_{reg}$ as
\begin{equation}
\mathcal{L}_{reg} = \mathbb{E} \, \| v_p - v\|^2_2 \,. 
\end{equation}

\subsubsection{Tag name classification}
As shown in Fig. \ref{fig:num}(c), a tag classifier (TGC) is introduced to ensure that \cz{the numerical representation retains the original knowledge of the tag name, using $h$ as the input}:
\begin{equation}
  \vspace{-2pt}
\mathcal{L}_{cls}= \mathbb{E} \, [ - y_{tag} \cdot \log \text{{\tt TGC}}(h)_{tag}] \,,
  \vspace{-1pt}
\end{equation}
where $y_{tag}$ is the ground truth label for the target numerical value's tag name,  and $\text{{\tt TGC}}(h)_{tag}$ denotes the probabilistic output of the TGC model for  label $y_{tag}$.
Note that this objective is optional, since sometimes there may be newly unseen tag names during the development of a specific field.

\subsubsection{Numerical contrastive learning}
To further strengthen the alignment degree between the numerical interval and the embedding distance, we propose a numerical contrastive loss $\mathcal{L}_{nc}$.
Concretely, given a target sample, traditional supervised contrastive learning  defines samples with the same label as positive and those with different labels (within the mini-batch) as negative. 
Differently, in our approach, we define the sample whose \textbf{numerical value is closest to the target value $v$ as positive and the rest as negative within each min-batch}.
Formally, given a numerical value $v$,
\begin{equation}
    \mathcal{L}_{nc}= \mathbb{E} \, [ 
    - log \frac{\exp \left(Sim\left({h}, {h}^{+}\right) / \tau\right)}{\sum_{{h}^{\prime} \in \mathcal{N}({v})} \exp \left(Sim\left({h}, {h}^{\prime}\right) / \tau\right)}
    ]\, ,
\end{equation}
where $\tau$ is the temperature hyper-parameter, $\mathcal{N}({v})$ is the in-batch negative embedding set, ${h}^{+}$ denotes the in-batch positive embedding and $Sim(\cdot , \cdot)$ is the cosine similarity function here.
\cz{This objective helps to smooth the numerical value changing process and stabilize the model (As shown in Fig. \ref{fig:vis})}. 

\subsubsection{Automatically weighted loss}
Considering that three training objective{s} are involved for numerical encoder learning, we exploit {an} automatically loss-weighted strategy \cite{DBLP:conf/cvpr/KendallGC18} for multi-task fusion, which weighs multiple loss functions by considering the homeostatic uncertainty of each task. 
This allows us to simultaneously learn various quantities with different units
or scales in both classification and regression settings.
Let $\mu_i$ be a learnable observation noise parameter toward task $i$  which captures how much noise contained in the outputs, we formulate the numerical loss function as follows:
$$
    \mathcal{L}_{num} = \frac{1}{2}
    (
    \frac{1}{\mu_1^2} \mathcal{L}_{reg} + 
    \frac{1}{\mu_2^2} \mathcal{L}_{cls} +
    \frac{1}{\mu_3^2} \mathcal{L}_{nc}
    ) + 
    \sum\nolimits_{i=1}^3 \log(1+\mu_i^2)\, .
$$

\subsubsection{Orthogonal regularization for Parameters}
\cz{In addition to the standard weight decay regularization applied to the entire network},
we introduce {an} orthogonal regularization for parameters in numeric transformation functions to mitigate the gradient explosion phenomenon when {the} network is deep \cite{DBLP:conf/iclr/BrockLRW17}.
This is achieved by adding the following term to the final loss function:
\begin{equation}
    \mathcal{L}_{num} \gets \mathcal{L}_{num} + 
    \lambda\sum\nolimits_{i=1}^{N}\left(\left\|{I}-{W}_v^{(i) \top} {W}_v^{(i)}\right\|_{F}^{2}\right)\, ,
\end{equation}
\cz{where $\lambda$ is a hyperparameter controlling the strength of the regularization and $I$ is the identity matrix.}

\subsection{Mask Reconstruction} \label{sec:mask}
Since firstly being proposed in BERT \cite{DBLP:conf/naacl/DevlinCLT19}, mask reconstruction  gradually becomes a general self-supervised pre-training strategy in large scale data pre-training \cite{DBLP:journals/corr/abs-1907-11692,DBLP:journals/taslp/CuiCLQY21,DBLP:conf/aaai/SunWLFTWW20}.

Not that the common objective for mask loss is:
\begin{equation}
\mathcal{L}_{mask} = \mathbb{E}\,[ - \sum\nolimits_{i=1}^{Len(w)} \log P(w_i\,|\,\mathcal{S}_{\backslash w})] \,,
\end{equation}
where $w$ represent{s} the target token sequence to be predict{ed} in the PLM's vocabulary,  $\mathcal{S}_{\backslash w}$ denotes the input sentence with  sub-sequence $w$ being masked, and the cross-entropy objective is adopt{ed} for masked token reconstruction in $P(w_i\,|\,\mathcal{S}_{\backslash w})$. 
In our work, we exclude those pre-defined prompt special tokens and numerical values from the candidate set of $w$.
The original data is wrapped by the prompt template, as introduced in Sec. \ref{sec:Unify}, to make up the candidates for {an} input sentence $\mathcal{S}$. 

\subsubsection{Masking Rate}
The masking rate refers to the proportion of masked tokens to the total number of tokens.
Most of the previous works follow the standard rate (i.e., $15\%$) in BERT. 
Nevertheless, recent research 
has shown that higher masking rates benefit training via adding discrepancy \cite{DBLP:journals/corr/abs-2202-08005}.
Thus, in this work we increase the masking rates from 15\% to 40\% during the KTeleBERT re-training.

\subsubsection{Masking Strategy}
We consider the following masking strategies in our work during the re-training stage:
\begin{itemize}
    \item The dynamic masking strategy in RoBERTa \cite{DBLP:journals/corr/abs-1907-11692} which dynamically changes the masking pattern applied to the training data in each step.
    \item The Chinese WWM strategy in MacBERT \cite{DBLP:journals/taslp/CuiCLQY21}. we use the LTP tool \cite{DBLP:conf/emnlp/CheFQ021} for Chinese whole-word split.
\end{itemize}

\subsection{Expert Knowledge Injection} \label{sec:kge}
\begin{figure}[htbp]
  \centering
  \vspace{-3pt}
  \includegraphics[width = 0.89\linewidth]{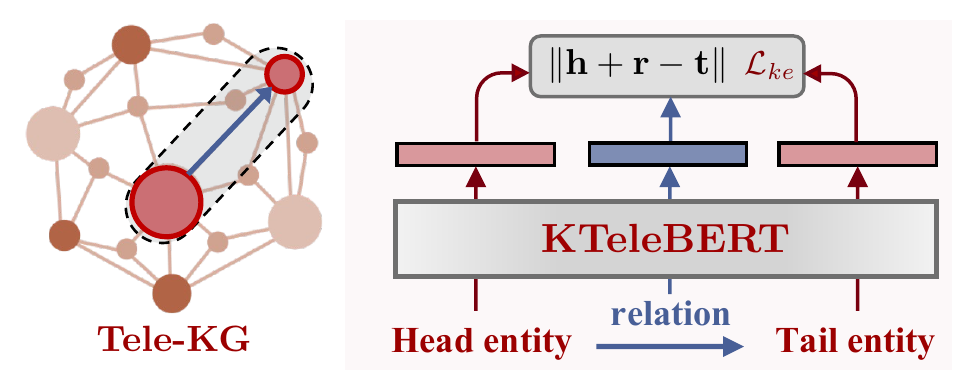}
  \caption{Text-enhanced KE progress in KTeleBERT.}
  \label{fig:kge}
  \vspace{-5pt}
\end{figure}

To enhance PLM's ability for explicit reasoning, 
we introduce a text-enhanced knowledge embedding (KE) objective for tele-expert knowledge injection, following the approach of KEPLER \cite{DBLP:journals/tacl/WangGZZLLT21}. 
As shown in Fig. \ref{fig:kge}, we warp entities and relations into textual sentences using the templates, and encode them using KTeleBERT to obtain their embeddings.

Let $e_{h}$, $e_{r}$, $e_{t}$ denote the embeddings of the given head entity, relation, and tail entity, respectively. We formulate the KE loss function $\mathcal{L}_{ke}$ as follows:
\begin{equation}
\begin{split}
    & \mathcal{L}_{ke}=-\log \sigma\left(\gamma-d_{r}(\mathbf{h}, \mathbf{t})\right) \\
    & -\sum_{i=1}^{n} p\left(h_{i}^{\prime}, r, t_{i}^{\prime}\right) \log  \sigma\left(d_{r}\left(\mathbf{h}_{i}^{\prime} \mathbf{t}_{i}^{\prime}\right)-\gamma\right)\, ,
\end{split}
\end{equation}
where $(h_{i}^{\prime}, r, t_{i}^{\prime})$ are negative samples, $\gamma$ is the margin, $\sigma$ is the sigmoid function.
Let $d_{r}$ be the TransE \cite{DBLP:conf/nips/BordesUGWY13} scoring function: 
\begin{equation}
    d_{r}(\mathbf{h}, \mathbf{t})=\|e_{h} + e_{r}-e_{t}\| \,.
\end{equation}
Note that we define the negative sampling policy as fixing the head entity and randomly sample a tail entity, and vice versa.

\subsection{Training Strategy for Multi-source Data} \label{sec:trainstrategy}
In order to achieve the co-training for KTeleBERT,
we define it as a multi-task learning (MTL) progress that combines different tasks across multi-source data.
Concretely, two strategies are considered to avoid forgetting the learned knowledge: 
\begin{itemize}
	\item Iterative multi-task learning (IMTL). We follow ERNIE2 \cite{DBLP:conf/aaai/SunWLFTWW20}, a continual multi-task pre-training framework from Baidu, to apply an iterative training method across tasks.
	\item Cooperative parallel multi-task learning (PMTL). The loss from different tasks are simply summed in each step. 
\end{itemize}

In addition, we also consider single-task learning (STL), i.e.,  using only the causal sentences and machine data for mask reconstruction. The details are recorded in Table \ref{tab:learnStrategy} where we unify the total training step of each task to 60k for comparison.

\begin{table*}[htbp]
\centering
\caption{Details about different learning strategy, including multi-task learning (MTL) and single task learning (STL). 
}
\begin{tabular}{c|c|ccc|c}
\toprule
\multirow{2}{*}{Strategy} & \multirow{2}{*}{Re-training task} & \multicolumn{3}{c|}{Training iterations (steps) } & \multirow{2}{*}{Training objective} \\ \cmidrule{3-5} 
& & Stage 1 & Stage 2 & Stage 3 &\\ 
\midrule
\multirow{1}{*}{Single-task Learning (STL)}  & Masking Reconstruction & \multicolumn{3}{c|}{60k} & $\mathcal{L}_{num}+\mathcal{L}_{mask}$ \\  
\midrule

\multirow{2}{*}{Parallel Multi-task Learning (PMTL)}  & Masking Reconstruction & \multicolumn{3}{c|}{60k} &  \multirow{2}{*}{$\mathcal{L}_{num}+\mathcal{L}_{mask}+\mathcal{L}_{ke}$}\\  \cmidrule{2-5}
& Knowledge Embedding & \multicolumn{3}{c|}{60k}  \\  
\midrule
\multirow{2}{*}{Iterative Multi-task Learning (IMTL)}  & Masking Reconstruction & 40k & 10k & 10k & $\mathcal{L}_{num}+\mathcal{L}_{mask}$ \\  \cmidrule{2-6}
& Knowledge Embedding & - & 40k & 20k & $\mathcal{L}_{ke}$ \\ 
	\bottomrule
\end{tabular}
\label{tab:learnStrategy}
  \vspace{-6pt}
\end{table*}

\section{Experiment}
In this section, we firstly \cz{provide implementation details   for pre-training and re-training, including the datasets,  experimental environment, and parameter settings.} 
Then, we validate TeleBERT and KTeleBERT on three tele-domain downstream tasks with MacBERT being used as a strong baseline,
demonstrating our model's superiority for fault analysis.
Further analysis for ANEnc is conducted to support our motivations.

{\subsection{Pre-training Details}}
\subsubsection{TeleBERT Pre-training}
\noindent\textbf{Datasets}: 
The Tele-Corpus  used for TeleBERT pre-training  includes various tele-domain data, such as the tele-question answering, software parameter descriptions, and daily maintenance cases, all sourced from the product documents. 
As outlined in Sec. \ref{sec:telebert}, data augmentation methods are used to generate a total $20,330$ k sentences (1.4GB).
\noindent\textbf{Environment}:
\cz{The pre-training of TeleBERT was performed on an $8$$\times$$8$ $32$G NVIDIA V100 cluster for $30$ epochs with a batch size of $4096$, for a total of $269$ hours.}

\subsubsection{KTeleBERT Re-training}
\noindent\textbf{Datasets}: 
As described in Sec. \ref{sec:Unify} and Sec. \ref{sec:kge}, the data for KTeleBERT re-training includes causal sentences, numeric-relate machine (log) data, and triples from the Tele-KG. 
To balance the data scale, we select $434$K causal sentences, 429K machine logs (alarms and KPI information), and 130K knowledge triples.
\noindent\textbf{Environment and Parameters}: 
For those coefficients, we set $\lambda$ to $1e$$-4$, temperature $\tau$ to $0.05$, margin $\gamma$ to $1.0$, learning rate (LR) to $4e$$-5$, and accumulation steps to $6$. Ten negative samples for each entity in $\mathcal{L}_{ke}$ are sampled.   
About 8 hours are spent to re-train KTeleBERT on four  24G NVIDIA RTX 3090 for 60K steps with a batch size (BS) of $256$. 

\subsubsection{\hyf{Service Delivery Paradigm}}
\label{sec:service-delivery}
Given a target name in tele-domain from related downstream tasks, we consider three types of data: 
\textbf{\textit{(\rmnum{1})}} ``only name'': pure literal name for the target. 
\textbf{\textit{(\rmnum{2})}} ``Entity mapping w/o Attr.'': the target name is mapped to an entity in Tele-KG by surface.
\textbf{\textit{(\rmnum{3})}} ``Entity mapping w/ Attr.'': The target name is mapped to an entity in Tele-KG with  attributes provided by downstream tasks appended.

Particularly, the data formats follow the basic templated rule shown in Fig. \ref{fig:KTeleBERT}, and our model encodes the content for those wrapped names with {\tt [CLS]} token's output embeddings  as the representations, which serve as the service embeddings for all fault analysis downstream tasks.

\subsection{Task 1: Root-Cause Analysis} \label{sec:task1}
\subsubsection{Task Description}
\cz{The root-cause analysis (RCA) task aims to identify the network element (NE) that is most likely to be the source of a fault in a tele-network. }
\cz{To accomplish this, we formulate it as a node ranking problem on a graph representation of the system network, where the nodes are NEs with edges representing the connections between them. By presenting a rank of all nodes, our model allows engineers to easily identify the real fault and consider alternative possibilities in case the output fault is not the correct one.}
We represent a tele-network as $\mathcal{G} = (\mathcal{V}, \mathcal{E}, \mathrm{X})$, where $\mathcal{V}$ is a set of nodes (i.e., NEs), $\mathcal{E}$ is a set of edges (i.e., connections between NEs), 
and \cz{$\mathrm{X} \in \mathbb{R}^{|\mathcal{V}| \times N}$ is the feature matrix with each row $i$ describing the abnormal event feature on corresponding node $v_{i}$  and $N$ being the total number of abnormal events.} 
For example, $x_{ij} = 3$ indicates that the abnormal event $j$ occurs three times on the network element $i$.
In practice, analysts typically collect information about a tele-network at a specific time slot when abnormal events occur, and we refer a tele-network in a time slot to a state. 
The goal of this task is to design a model $f$ that can map a state of a tele-network to a score vector of nodes, $s = f(\mathcal{G})$, where $s \in \mathbb{R}^{|\mathcal{V}|}$  representing the scores of nodes.
Note that the higher the score, the more likely it is that the corresponding node is the root cause.

\subsubsection{Method}
\begin{figure}[t]
\centering
\includegraphics[width = 0.9\linewidth]{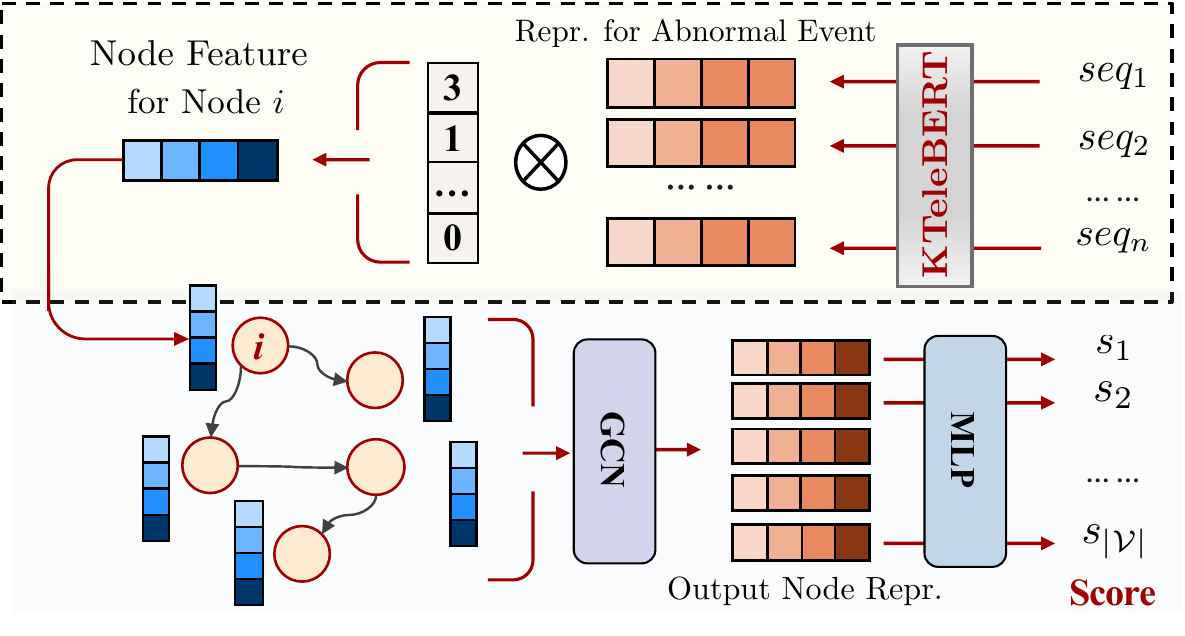}
\vspace{-2pt}
\caption{Method overview for root-cause analysis.}
\label{fig:root-cause-method}
\vspace{-8pt}
\end{figure}
The method for handling RCA is mainly based on our KTeleBERT and the Graph Convolutional Networks (GCNs) \cite{GCN}, as shown in Figure \ref{fig:root-cause-method}.
First, for a graph $\mathcal{G}$, we use the encoder (e.g., KTeleBERT) to obtain the representations of abnormal events 
\cz{by using the sequence format described in Section \ref{sec:service-delivery} as the input}: 
\begin{equation}
\vspace{-1pt}
    \mathbf{E}_i = {\rm KTeleBERT}(seq_i),
\vspace{-1pt}
\label{eq:root-cause-encode-event}
\end{equation}
where $seq_i$ is the input sequence for the $i$-th abnormal event.
$\mathbf{E}_i \in \mathbb{R}^{1 \times d}$ is the representation vector for abnormal event $i$, and $\mathbf{E} \in \mathbb{R}^{n \times d}$ is the matrix for all abnormal events with $d$ as the vector dimension.
\cz{We use these representations to initialize the node embeddings based on the node feature matrix $\mathbf{X}$. 
Specifically, 
the input node representations $\mathbf{H} \in \mathbb{R}^{|\mathcal{V}| \times d}$ for GCN are initialized via 
weighted average pooling:
}
\begin{equation}
\vspace{-1pt}
    \mathbf{H}_j = \frac{\mathbf{x}_{j} \mathbf{E}}{\sum{\mathbf{x}_{j}}}\, ,
\vspace{-1pt}
\end{equation}
where $\mathbf{x}_{j} \in \mathbb{R}^{1 \times n}$ is 
a vector indicating how many times an abnormal event happens on node $j$. 

\cz{Next, the node representation is updated throughout each of the GCN layer:} 
\begin{equation}
\mathbf{H}^{l}=\sigma\left(\widetilde{\mathbf{D}}^{-\frac{1}{2}} \widetilde{\mathbf{A}} \widetilde{\mathbf{D}}^{-\frac{1}{2}} \mathbf{H}^{l-1} \Omega^{l}\right) \, ,
\end{equation}
where $\mathbf{A}$ is the adjacency matrix of a graph $\mathcal{G}$,  $\widetilde{\mathbf{A}} = \mathbf{A} + \mathbf{I}$ is the adjacency matrix with self-loop, $\widetilde{\mathbf{D}}$ is the degree matrix of $\widetilde{\mathbf{A}}$, $\sigma$ is an activation function, $\Omega$ is a layer-specific trainable parameter. More precisely, $\mathbf{H}^{0} = \mathbf{H}$ and $\mathbf{H}^{L}$ is the output node representations.
Finally, the output node representations are passed through a 2-layer multi-layer perceptron (MLP) to calculate the final score for each node \cz{that represents the likelihood of each node being the root cause}:
\begin{equation}
    s = f_{s}(\mathbf{H}^{L}) \, .
\end{equation}
We use the logistic loss to train our model, where the labeled root cause nodes are treated as positive ($y=1$) and others as negative ($y=-1$). The parameters in GCN and MLP are optimized by minimizing the following objective:
\begin{equation}
    \mathcal{L}_{rca} = \sum_{\mathcal{G}_i} \sum_{j \in \mathcal{V}_i} \log(1+\exp(-y_{j}s_j)) \,.
\end{equation}

\subsubsection{Evaluation} 
\cz{We evaluate our method for RCA using a dataset containing different states of the tele-network, each with a labeled root cause.
The data statistics are summarized in Table \ref{tab:root-cause-data}, including the number of graphs, the number of graphs features, and the average number of nodes/edges in this dataset.}
\cz{
Besides, we also compare our method with the ``Random'', which uses random valued vectors drawn from a uniform distribution to represent the abnormal events.}
\begin{table}[!htbp]
\centering  
\caption{Data statistics for root-cause analysis.}
\centering
\begin{tabular}{cccc}
\toprule
\# Graph & \# Feature & \# Node & \# Edge \\
\midrule
127 & 349 & 10.96 & 51.15 \\
\bottomrule
\end{tabular}
\label{tab:root-cause-data}
\vspace{-4pt}
\end{table}
\begin{table}[!htbp]
\centering  
\vspace{-1pt}
\caption{Evaluation results for root-cause analysis.}
\centering
\begin{tabular}{lcccc}
\toprule
Method & MR $\downarrow$ & Hits$@$1 & Hits$@$3 & Hits$@$5 \\
\midrule
Random & 2.47 & 54.88 & 75.00 & 88.67  \\
MacBERT & 2.16 & 59.64 & 82.68 & 90.85  \\
\midrule
TeleBERT & 2.09 & 62.65 & 83.52 & 92.46  \\
\cdashline{1-5}[4pt/1pt]
KTeleBERT-STL & 2.06 & 63.66 & 83.21 & 91.87  \\
~~~~ w/o ANEnc & 2.13 & 60.72 & 82.96 & 90.80 \\
\midrule
KTeleBERT-PMTL & 2.03 & \textbf{65.96} & 84.98 & \textbf{92.63}  \\
KTeleBERT-IMTL & \textbf{2.02} & 64.78 & \textbf{85.65} & 91.13  \\
\bottomrule
\end{tabular}
\label{tab:root-cause-res}
\vspace{-4pt}
\end{table}

\ul{Implementation Details.} 
We unify the default dimension of representations to $768$, and use 2-layers GCNs with the hidden dimensions of $1024$ and the output dimensions of $512$.
\cz{The MLP used to transform output representations from GCNs to scores has 2 layers with 128-dimensional hidden layers.}

\cz{We use the K-fold validation to evaluate the models.} Specifically, 
\cz{the dataset is split into 5 folds, with 1 fold used as the testing set, 1 fold as the validation set, and the remaining folds used as the training set. The results are reported as the average of the 5 folds.} 
While for metrics, we use the mean rank (i.e., mean rank of labeled root causes on the node ranking lists by predicted scores in graphs) and Hits at N (Hits$@$N) (i.e., the proportion of labeled root causes ranked in top N in graphs ) to evaluate our model.

\ul{Results Analysis.} The results in Table \ref{tab:root-cause-res} 
\cz{indicates that using abnormal event representations from KTeleBERT could achieve better performance. 
Specifically, the best Hits$@$1 of KTeleBERT obtains $5.28\%$ relative improvements compared with TeleBERT, that is $10.60\%$ compared with MacBERT. 
The pool performance for ``Random'' supports the existence of upstream tele-domain pre-training for downstream task gain.
}

\subsection{{Task 2: }Event Association Prediction} \label{sec:task2}
\subsubsection{Motivation and Problem Formulation}
\cz{We \gyx{develop} the event association prediction (EAP) task \gyx{by} representing each event in a low-dimensional vector (i.e. event embedding) and learning the associations between events based on the embedding computation.
To achieve this, we propose a \textit{trigger} relation-specific space \gyx{where} events are \gyx{embedded} and their similarities are measured to predict the trigger relationships between event pairs.}
Formally, let $e_i$ and $e_j$ be a pair of input events, their similarity score $s_{ij}$ is $s_{ij} = f(\bm{e}_i, \bm{e}_j)$,
where $\bm{e}_i$ and $\bm{e}_j$ are the vector representations of the pair of events, and $f$ represents the similarity measurement function.
If there exists a trigger relationship between $e_i$ and $e_j$, the embeddings are similar in the \textit{trigger} relation space, otherwise they are dissimilar.
\cz{To \gyx{better learn the vectors representations of these} events, we \gyx{explore and embed some valuable information that characters them as the event initialization, instead of random initialization. The introduced information includes:}
}
 \begin{itemize}
  \vspace{-4pt}
     \item its literal name, which shortly and abstractly describes the event in text, and reveals some fault patterns such as by word co-occurrence;
     \item the topological environment of the network element it depends on, in general, a fault event is generated from a network element and the topological connections between the network elements decide the information flow in a network, that is to say, two events whose network elements that are adjacent are more likely to have the trigger relationship;
     \item its machine data, which reflects the running context that causes the fault event, such as its occurrence time.
  \vspace{-1pt}
 \end{itemize}

 \subsubsection{Method}
 \begin{figure}[htbp]
  \centering
  \vspace{-2pt}
  \includegraphics[width = 0.98\linewidth]{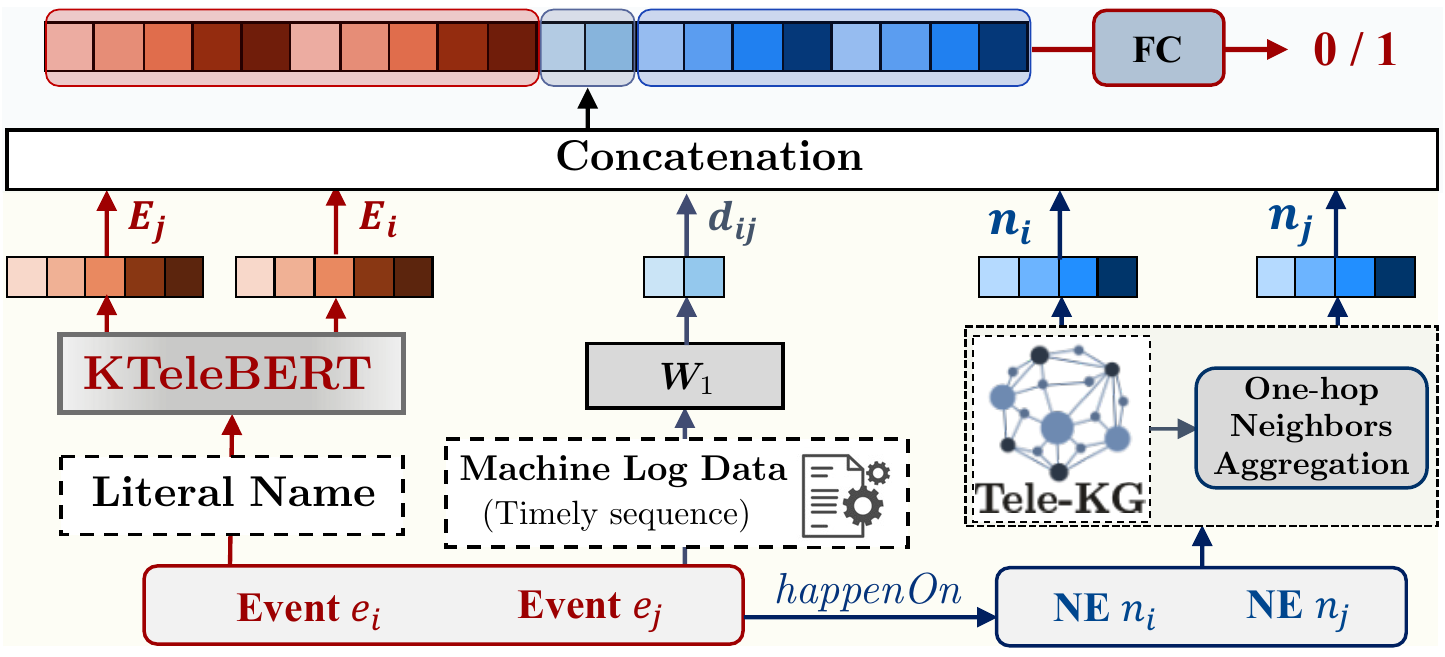}
  \caption{Method overview for event association prediction.}
  \label{fig:eventFramework}
   \vspace{-6pt}
\end{figure}
\cz{We \gyx{apply different} strategies to embed the different types of information \gyx{listed above.} The overview of the method can be found in Fig. ~\ref{fig:eventFramework}.}
 Specifically, given a target event pair ($e_i$, $e_j$), we first embed their corresponding literal names $\textit{seq}_i$ and $\textit{seq}_j$ in a way analogous to \cz{the RCA task, i.e., according to Eq. (\ref{eq:root-cause-encode-event}),}
 we have 
 $\mathbf{E}_i = {\rm KTeleBERT}(\textit{seq}_i)$ and $\mathbf{E}_j = {\rm KTeleBERT}(\textit{seq}_j)$.
 Next, to encode the topological environment, \cz{for the NE $n_i$ and $n_j$ on which these two events depend, we aggregate their one-hop neighbors in the network graphs.} Formally, for $n_i$, we have
\begin{equation}
 \bm{n}_i = \frac{1}{\mathcal{N}_i}\sum\nolimits_{k \in \mathcal{N}_i} \bm{n}_k \, ,
\end{equation}
 where $\bm{n}_i$  denotes its aggregated embedding and $\mathcal{N}_i$ is its one-hop neighbor set including itself. 
\cz{Furthermore, 
we also \gyx{explore events'} temporal aspect \gyx{from the machine data,} and \gyx{try to} encode the time difference \gyx{to reveal} the sequential features between the events.} 
Formally, for the occurrence time $t_i$ and $t_j$ of $e_i$ and $e_j$, we first compute the D-value and pass it into a FC network with a weight parameter of $\bm{W}_1$, as
\begin{equation}
 \bm{d}_{ij} = \bm{W}_1(t_i-t_j) \, .
\end{equation}
\cz{The motivation 
behind encoding the occurrence time difference 
is built upon our observations. That is, 
the \gyx{\textit{trigger}} relationship between \gyx{two} events that occur \gyx{at a long margin} may be \gyx{weak}, while \gyx{two} events that occur \gyx{nearly} or almost simultaneously may have relationships beyond \gyx{\textit{trigger}}, such as \gyx{\textit{co-occurrence}}. 
Therefore, we leverage a FC layer with weight parameter $\bm{W}_1$ to automatically determine the reasonable arising time interval for the current event relatedness.}

Then, we concatenate these vectors to generate a final representation for the input event pair and pass it into another FC network with weight parameter $\bm{W}_2$ to predict the similarity score with $[\cdot ; \cdot]$ referring to the concatenation operation:
\begin{equation}
 s_{ij} = \bm{W}_2 [\mathbf{E}_i; \mathbf{E}_j; \bm{n}_i;  \bm{n}_j; \bm{d}_{ij}] \, .
\end{equation}

\cz{A standard binary cross-entropy loss with the softmax activation function ($\sigma$) is minimized to train our model for all event pairs in the datasets:} 
\begin{equation}
\begin{aligned}
\mathcal{L}_{eap} = -\frac{1}{|\mathcal{P} \wedge \mathcal{P}'|} \sum_{(e_i, e_j) \in \mathcal{P} \wedge \mathcal{P}'} y_{ij} \cdot log(\sigma (s_{ij})) \\
    + (1-y_{ij}) \cdot log(1-\sigma(s_{ij})) \, ,
\end{aligned}
\end{equation}
where $\mathcal{P}$ is the set of positive event pairs, i.e., there exists a trigger relationship between events in the pair. For each positive pair, we randomly replace one of the two events with other events to constitute the negative set $\mathcal{P}'$, requiring that none of them exists in current positive set.
$y_{ij}$ is the label of the given event pair ($e_i$, $e_j$), whose value is 1 when the pair is positive and 0 otherwise.
In prediction, 
\cz{the candidate event pair is input to the model for  similarity score computation.}
\subsubsection{Evaluation} 

\begin{table}[ht]
\centering  
\caption{Data statistics for event association prediction.}
\centering
\begin{tabular}{ccccc}
\toprule
\# Events &  \makecell[c]{\# Event Pairs \\ (positive)} &  \makecell[c]{\# Event Pairs \\ (negative)} & \makecell[c]{\# MDAF \\ packages} & \makecell[c]{\#  Nework \\ Elements}
\\
\midrule
86 & 2141 & 2141 & 104 & 31 \\
\bottomrule
\end{tabular}
\label{tab:eventData}
\vspace{-2pt}
\end{table}
\begin{table}[ht]
\centering  
\caption{Evaluation results for event association prediction.}
\centering
\begin{tabular}{lcccc}
\toprule
Methods & Accuracy & Precision & Recall & F1-score \\
\midrule
Word Embeddings & 64.9 & 66.4 & 96.8 & 78.7 \\
MacBERT  & 64.3 & 65.9 & 96.1 & 78.2   \\
\midrule
TeleBERT  & 70.4 &	71.4 &	95.1 &	81.5\\
\cdashline{1-5}[4pt/1pt]
KTeleBERT-STL &  \textbf{77.3} &	\textbf{76.6} &	96.6 &	\textbf{85.4} \\
~~~~ w/o ANEnc & {76.0} & {76.1} & {95.1} & {84.5} \\
\hline
KTeleBERT-PMTL & {68.5} & {68.8} & \textbf{99.1} & {81.3} \\
KTeleBERT-IMTL & {73.5} & 73.8 & {95.6} & 83.2 \\
\bottomrule
\end{tabular}
\label{tab:eventResults}
\vspace{-8pt}
\end{table}

\cz{To evaluate the performance of our method, we use a dataset consisting of event pairs that are known to have trigger relationships and have been validated by tele-experts.
We split the dataset into two disjoint sets: $80\%$ for training and $20\%$ for testing.}
For each event pair, we collect MDAF packages to provide the  machine data and a graph of NEs to provide their topological environment.
The detailed statistics are shown in Table~\ref{tab:eventData}. \cz{Note that we further use learnable word embeddings \gyx{to represent the literal names of events} as \gyx{another baseline}, where a name sequence \gyx{will be} separated into multiple words, each of which is randomly initialized with a 768-dimensional vector and the events are represented by averaging their word embeddings.}

\ul{Implementation Details.}
\cz{The learning rate is set to $0.01$ and the batch size is set to $32$.
We set the shapes of parameter matrices 
to $\bm{W}_1 \in \mathbb{R}^{1 \times 2}$ and 
$\bm{W}_2 \in \mathbb{R}^{540 \times 2}$.} 
We also perform random data splits for 5-fold cross-validation to ensure robust results.
Since the task is modeled as a binary classification task, we report the evaluation results with widely used metrics of the accuracy, precision, recall, and F1-score.

\ul{Result Analysis.} 
The results compared with baselines are shown in Table~\ref{tab:eventResults}.
\cz{We observe that the representations of event literal names by our TeleBERT or KTeleBERT perform better than those generated by the solutions raised as baselines.
Additionally, the results indicate that domain-specific methods such as word embeddings learned by known event pairs and TeleBERT trained using Tele-Corpus perform better than general-purpose models such as MacBERT. 
Overall, in most situations, our models show their superiority in encoding event literal information and well support the EAP task.

}

\subsection{Task 3: Fault Chain Tracing} \label{sec:task3}
\vspace{-1pt}
\subsubsection{Task Description}
\cz{The task of Fault Chain Tracing (FCT) involves identifying the root cause of a fault within a network by completing the fault propagation chain. Given a tele-network of fault alarms, the objective is to connect those faults in the correct sequence to form complete fault chains, which can then be used to trace the origin fault. It requires a thorough understanding of the network's topology and the relationships between different fault alarms, as well as the ability to identify patterns and connections that may not be immediately obvious.}
\vspace{-5pt}
\begin{figure}[htbp]
  \centering
  \includegraphics[width = 1.0\linewidth]{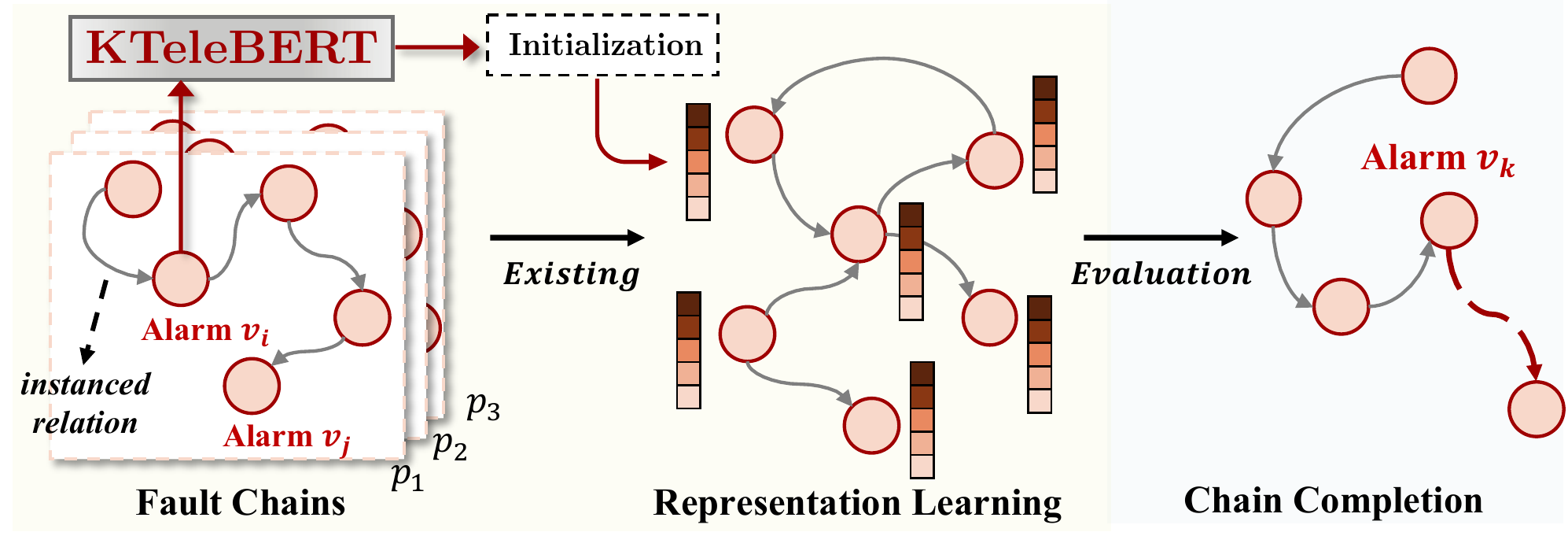}
  \caption{Method overview for fault chain tracing.}
  \label{fig:faultchain}
   \vspace{-8pt}
\end{figure}

\subsubsection{Problem Formulation}
Since fault chains are composed of a series of correlative triples, we formulate the the FCT problem as the link prediction task in the fault chain.
The tele-network can be represented by the heterogeneous graph $\mathcal{G} = (\mathcal{V}, \mathcal{E}, Q, \mathcal{P})$, where $\mathcal{V}$ is a set of nodes (alarms) and
\cz{$\mathcal{E}$ is a set of edges (relations between the alarm NE instances).
A fact set, $Q$, with quadruples of $(h,r,t,s)$ representing a connection between the nodes (alarms) $h$ (head) and $t$ (tail) with relation $r$ and a probability score $s$.
Note that $h, t \in  \mathcal{V}$, $r \in \mathcal{E}$, $s \in \mathbb{R}_{[0,1]}$.}
A higher $s$ means it is more likely for $h$ and $t$ to be connected by $r$.
$\mathcal{P} = \{p_1, p_2, ... \}$ denotes the fault propagation path (fault chain), which consists of a set of alarms.
\cz{The fault chains in real tele-scenario are sometimes incomplete, thus we need to analyze the associations among candidate alarm nodes in the range of one-hop or two-hop steps.}
Note that the heterogeneous graph $\mathcal{G}$ is constructed by the paths in $\mathcal{P}$.
\cz{Our goal is to design a model $f$ for fault chain completion}, i.e., $\mathcal{P'} = f(\mathcal{G})$ where $\mathcal{P'}$ is the completed path.

\subsubsection{Method}
Our method is \cz{composed of the following three steps (Fig. \ref{fig:faultchain}): }
\textbf{\textit{(\rmnum{1})}} \ul{Rules Lightning}.
\cz{In real-world scenarios the tele-networks have complex fault structures. So, given the graph $\mathcal{G}$ we need to filter out irrelevant alarms and NEs using the pre-defined rules to obtain a filtered graph $\mathcal{G'}$.}
For example, we use rules like $(v_i, cause, v_j)$ to get the relevant alarms and their corresponding NEs.
\textbf{\textit{(\rmnum{2})}} \ul{Initialization of Pre-training Knowledge}. 
\cz{Following Eq. (\ref{eq:root-cause-encode-event}), we use the encoding model (e.g., KTeleBERT) to obtain informative embeddings for each node in the filtered graph $\mathcal{G'}$, which is an important step for capturing the implicit associations between alarms and NEs.} 
\textbf{\textit{(\rmnum{3})}} \ul{Training and Prediction}. 
\cz{We leverage a generalizing translation-based method for uncertain knowledge graph embedding to model the probabilistic knowledge in the graph.}
We follow the paradigm in \cite{GTransE} for probabilistic knowledge representation learning, using an objective that takes into account the confidence of each quadruple:
\begin{small}
\begin{equation}
    \mathcal{L}_{fct} =  \sum_{(h,r,t,s) \in Q}{ \sum_{(h',r,t',s) \in Q'}{[d_{r}(h, r, t) - d_{r}(h', r, t')
     + s^{\alpha}M]_{+}} } \, ,
\end{equation} 
\end{small}
\cz{where $d_{r}$ is the score function for $(h, r, t)$ toward the fact quadruple $q$ with $s$ as the confidence, 
$M$ is the margin hype-parameter, 
and $\alpha$ is an adjusting hyper-parameter.} 
Then our framework is able to predict the missing links in incomplete paths through self-supervised representation learning.

\subsubsection{Evaluation} 
\cz{To evaluate the performance gain from the upstream model on our algorithm, we create a dataset of incomplete fault chain paths by masking those first-hop relations between alarms. The dataset is split into three sets: training, validation and testing, with the statistics presented in Table~\ref{tab:fault-chain-traceability-data}. Additionally, we use those randomly initialized embeddings of entities and relations in the knowledge graph as a basic encoding  for comparison.
The evaluation metrics used in this task are MRR, Hits$@$1, Hits$@$3, and Hits$@$10. 
} 
\begin{table}[ht]
\centering  
\vspace{-4pt}
\caption{Data statistics for fault chain tracing.}
\centering
\begin{tabular}{ccccc}
\toprule
\# Nodes & \# Edges & \# Train & \# Valid & \# Test \\
\midrule
243 & 100 & 232 & 33 & 32 \\
\bottomrule
\end{tabular}
\label{tab:fault-chain-traceability-data}
\vspace{-3pt}
\end{table}

\begin{table}[ht]
\centering  
\caption{Evaluation results for fault chain tracing.}
\centering
\begin{tabular}{lcccc}
\toprule
Method & MRR & Hits$@$1 & Hits$@$3 & Hits$@$10 \\
\midrule
Random & 58.2 & 56.2 & 56.2 & 62.5  \\
MacBERT & 65.9 & 62.5 & 65.6 & 68.8  \\
\midrule
TeleBERT & 69.0 & 65.6 & 71.9 & 71.9  \\
\cdashline{1-5}[4pt/1pt]
KTeleBERT-STL & {73.6} & {71.9} & {71.9} & {78.1} \\
~~~~ w/o ANEnc & {67.5} & {65.6} & {65.6} & {71.9} \\
\hline
KTeleBERT-PMTL & {87.3} & {84.4} & {87.5} & {93.8} \\
KTeleBERT-IMTL & \textbf{94.8} & \textbf{93.8} & \textbf{93.8} & \textbf{100.0} \\
\bottomrule
\end{tabular}
\label{tab:fault-chain-traceability}
\vspace{-6pt}
\end{table}


\ul{Implementation Details}.
\cz{We set the batch size and the number of negative samples as \{1024, 1000\}.
The LR is set as ${10^{-5}}$ with $2000$ as the hidden embedding size.}

\ul{Results Analysis}.
\cz{The results presented in Table \ref{tab:fault-chain-traceability} show that our KTeleBERT models achieve the best performance and a significant increase compared to the baseline methods. 
Specifically, the results of our KTeleBERT-IMTL model performs the best, with $94.8$ in MRR, $93.8$ in Hits$@$1, $93.8$ in Hits$@$3, and $100.0$ in Hits$@$10.}

\subsection{Discussion on  Adaptive Numeric Encoder} \label{exp:ablation}

\subsubsection{Numerical Contrastive Learning}
\begin{figure}[htbp]
  \centering
  \vspace{-2pt}
  \includegraphics[width = 0.95\linewidth]{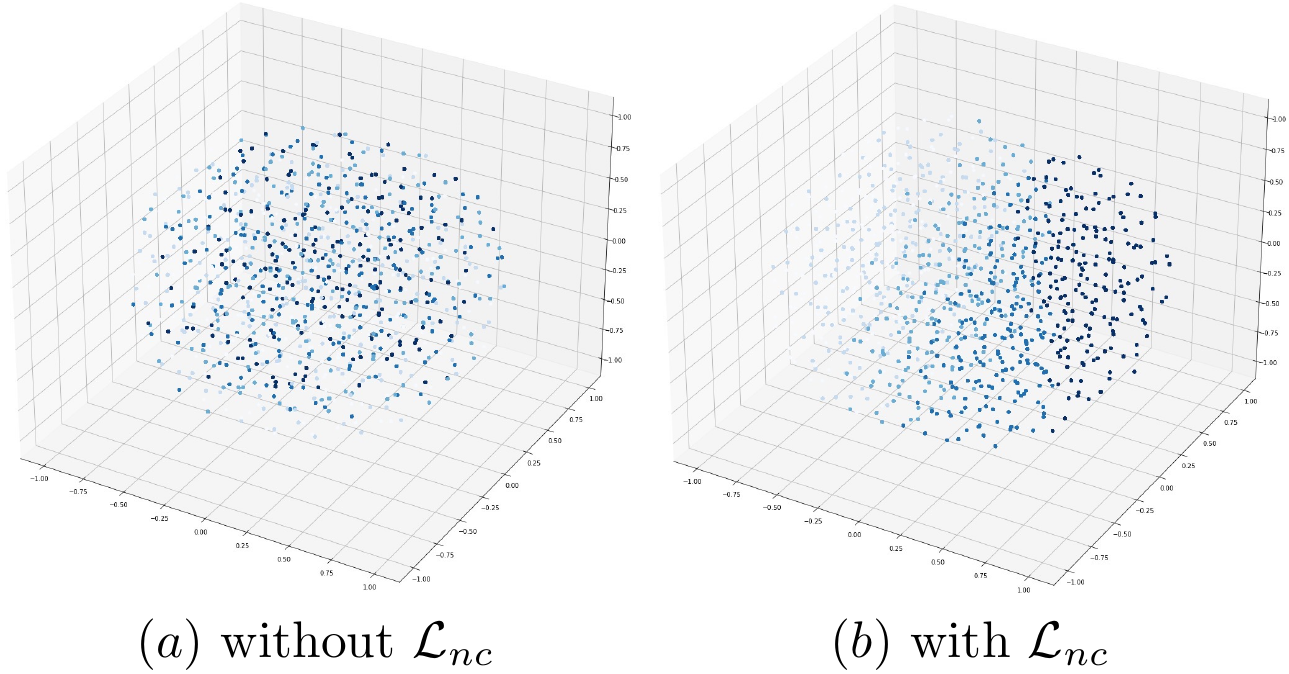}
  \caption{Visualization for numerical value embedding.
  }
  \label{fig:vis}
  \vspace{-6pt}
\end{figure}

\cz{
We visualize the numerical value embedding generated by ANEnc (with or without the objective $\mathcal{L}_{nc}$) through dimension reduction, as shown in Fig. \ref{fig:vis}(a) and Fig. \ref{fig:vis}(b), where a lighter color represents a smaller value.
We observe that the continuous changes among values from small to large are effectively mapped into the (3-D) vector spaces with $\mathcal{L}_{nc}$ applied, clearly proving that our proposed numerical contrastive learning 
 strategy allows the model to effectively understand the magnitude relationships among various numerical values.
}

\subsubsection{Adaptive Numeric Encoder} \label{sec:newANE}
As the result shown in Table \ref{tab:root-cause-res}, \ref{tab:eventResults}, 
and \ref{tab:fault-chain-traceability}, our KTeleBERT not only exceed\hyf{s} the simple baselines like the random embedding initialization and the learnable word embedding, but also perform\hyf{s} better than the strong baseline like the TeleBERT and the MacBERT.
We find that 
\cz{our adaptive numeric encoder (ANEnc) module 
play a crucial role in this process, which
consistently improves the performance in multiple downstream tasks when using the STL version of KTeleBERT as the control model.} 

\cz{
\begin{figure}[!htbp]
\vspace{-2pt}
  \centering
  \vspace{-3pt}
  \includegraphics[width = 0.88\linewidth]{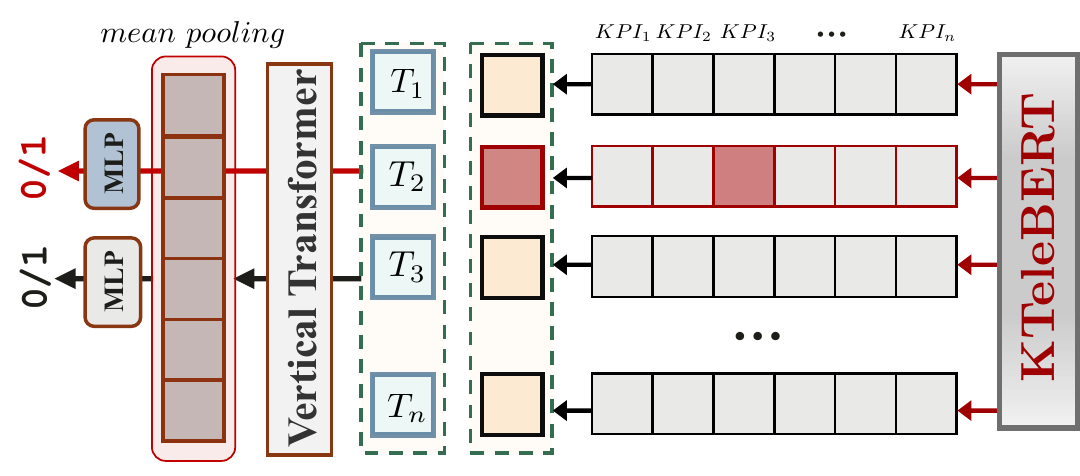}
  \vspace{-2pt}
  \caption{Method overview for abnormal KPI detection.}
  \label{fig:admodel}
   \vspace{-9pt}
\end{figure}
To further justify the advantages of ANEnc on tackling numerical data, we introduce a new task: \textbf{Abnormal KPI Detection}, which aims to discover the abnormal value change within the real-time KPI flow of the tele-system. 
We emphasize that multiple KPI indicators are recorded simultaneously on each NE (device) at regular intervals, as shown in Fig. \ref{fig:TeleKG}(a). This requires the model to encode not only the separate machine (horizontal) data point but also those continuous machine data (vertical) segments, while understanding the specific and inconsistent change frequencies and rules associated with each KPI.
Concretely, we design an vertical transformer \cite{DBLP:conf/acl/YinNYR20} module (2 layers 3 headers) with learnable position embedding attached to model the temporal relationship. Two objectives are involved to train the entire model:
\begin{align}
\vspace{-2pt}
\hat{\mathcal{L}}_{ad} & =\operatorname{BCE}\left(p\left(\hat{y} \mid x_1, x_2, \dots ,x_n \right)\right) \,, \\
\mathcal{L}_{ad}^{n} & =\operatorname{BCE}\left(p\left(y_n \mid x_n \right)\right) \,,
\vspace{-2pt}
\end{align}
where $\hat{\mathcal{L}}_{ad}$ denotes the objective toward the overall state ($\hat{y} \in \{0,1\}$) of the machine data segment and $\mathcal{L}_{ad}^{n}$ targets at the separate state for machine data point ($y_n \in \{0,1\}$) of a certain moment. The machine data embedding,  encoded by a basic machine data encoder, is denoted as $x_n$. 
Note that the encoder is frozen during training and we use KTeleBERT-STL as the encoder to independently demonstrate the effectiveness of our ANEnc module.
The overall loss is:
\begin{equation}
    \mathcal{L}_{AD} =\hat{\mathcal{L}}_{ad} + \frac{1}{N_{seg}}\sum\nolimits_{i=1}^{N_{seg}} \mathcal{L}_{ad}^{i} \,, 
\end{equation}
where $N_{seg}$ is the length of machine data segment.
For experiment, we collect $12,347$ temporal data segment ($98,776$ data point) in a real tele-network to constitute our datasets, as shown in Table \ref{tab:addata}. 
Note that a segment is considered abnormal when it contains at least one abnormal point.
\begin{table}[!htbp]
\centering  
\vspace{-4pt}
\caption{Data statistics for abnormal KPI detection.}
\centering
\begin{tabular}{cccc}
\toprule
\# Abnormal Seg.  & \# Abnormal Poi. & \# Normal Seg. & \# Normal Poi. \\
\midrule
4,510 & 7,512 & 7,864 & 90,912  \\
\bottomrule
\end{tabular}
\label{tab:addata}
\vspace{-6pt}
\end{table}
\begin{table}[!htbp]
\centering
\vspace{-2pt}
\caption{Evaluation results for abnormal KPI detection.}
\centering
\begin{tabular}{ccc}
\toprule
Method & Recall Poi. ($\%$) & Recall Seg. ($\%$)  \\
\midrule
KTeleBERT (w/o ANEnc) & {0.00} & {70.79}   \\
KTeleBERT (w/ ANEnc) & \textbf{56.57} & \textbf{91.48}  \\
\bottomrule
\end{tabular}
\label{tab:adres}
\vspace{-6pt}
\end{table}

We apply the focal loss strategy \cite{DBLP:journals/pami/LinGGHD20} to alleviate the label imbalance problem since the abnormal machine data point is quite rare compared to the normal ones. The results about abnormal data recall are shown in Table \ref{tab:adres}, where significant improvement ($\uparrow$$56.57\%$ in abnormal data point recall) is achieved with ANEnc adopted in, demonstrating the superiority of our ANEnc module in capturing small changes in 
numerical values and their internal connections.
Furthermore, we visualize those machine data embedding (before the MLP layer) in vector space to compare the spatial distribution of normal and abnormal machine log data, as shown in Fig. \ref{fig:kpivis}, finding that with our ANEnc module, the abnormal data is reflected in the vector space in the form of outliers and the model's sensitivity to abnormal data is greatly enhanced. 

\begin{figure}[!htbp]
  \centering
  \vspace{-3pt}
  \includegraphics[width = 0.82\linewidth]{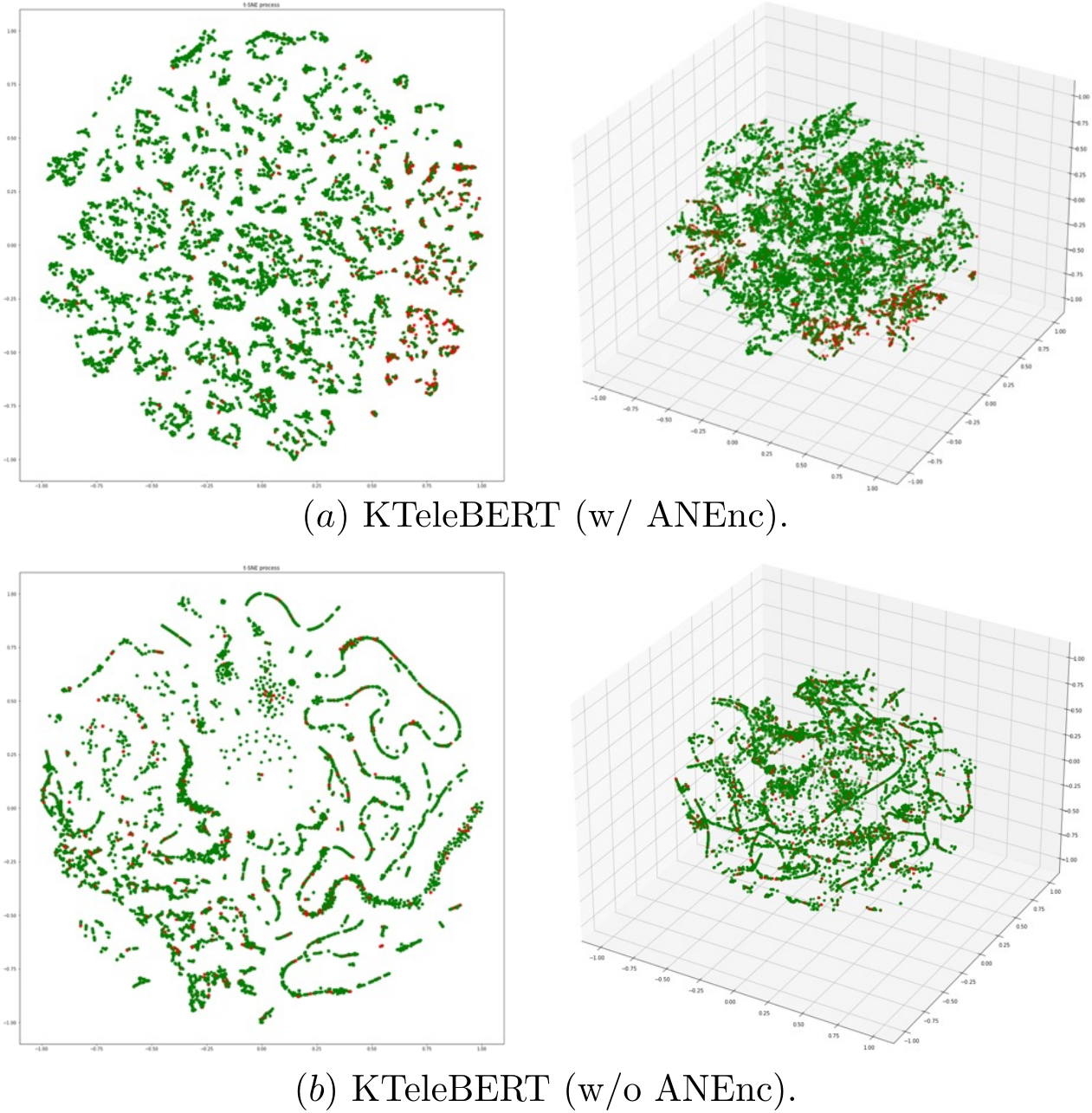}
    \vspace{-2pt}
  \caption{Visualization for machine data embedding in abnormal KPI detection task with the {\color{red} red} referring to the anomaly.}
  \label{fig:kpivis}
   \vspace{-6pt}
\end{figure}

}

\section{Related Work} 
\subsection{Pre-trained Language Model}
\cz{Recent advancements in PLMs have led to significant innovations in NLP communities.}  
Works like ERNIE  \cite{DBLP:journals/corr/abs-1904-09223}, spanBERT \cite{DBLP:journals/tacl/JoshiCLWZL20} and StructBERT  \cite{DBLP:conf/iclr/0225BYWXBPS20} have extended BERT  \cite{DBLP:conf/naacl/DevlinCLT19} by incorporating novel token-level and sentence-level pre-training tasks.
\cz{In addition to pre-training on sentence-like corpora,}
some researchers focus on injecting structured data (e.g., KGs) into PLMs for explicit knowledge integration. 
Specifically, KEPLER \cite{DBLP:journals/tacl/WangGZZLLT21} 
 integrates a text-enhanced knowledge embedding (KE) module without modifying the model structure. 
 KG-BERT \cite{DBLP:journals/corr/abs-1909-03193} 
 \cz{treats KG triples as textual sequences, using entity/relation descriptions as input.} 
K-BERT \cite{DBLP:conf/aaai/LiuZ0WJD020} 
employs soft positions and visible matrices to limit the knowledge scope.
Moreover, a series of table pre-training frameworks are also proposed, following the typical NLP pre-training paradigms. 
TaBERT \cite{DBLP:conf/acl/YinNYR20}  combines a row-wise transformer with column-wise vertical attention layers for handling  the hierarchical  structure of tabular data.
TableFormer \cite{DBLP:conf/acl/YangGUHGP22} introduces a structurally aware table-text encoding architecture, 
\cz{incorporating tabular structural information through learnable attention biases.} 
To enable 
numeric learning, Tapas \cite{DBLP:conf/acl/HerzigNMPE20} devises the rank embedding  for column-wise number comparison, 
and TUTA  \cite{DBLP:conf/kdd/WangDJLFHZ21} distinguishes those numerical values via embedding over various discrete numerical features.
However, the fine-grained numerical information encoding is rarely studied in depth in these works,
 making it challenging for models to analyze the relationships among close values.
\subsection{Numerical Information Encoding}
Existing numerical learning methods \cite{DBLP:conf/ijcai/GuoTYLH17,DBLP:conf/cikm/SongS0DX0T19,DBLP:conf/kdd/GuoCTZLH21} 
mainly focus on learning independent field features 
to distinguish different numerical meanings where the number of field is limited.
Besides, there exist works in KG community that define the numerical value encoding for attribute values as a n-gram encoding \cite{DBLP:conf/aaai/TrisedyaQZ19} problem or use a convolutional neural network (CNN) to extract features \cite{DBLP:conf/ijcai/ZhangSHCGQ19}. 
However, they seldom consider the fine-grained encoding for numerical data, making it hard to analyze the relationships among values from different fields.

\cz{
\subsection{Log-based Anomaly Detection} \label{sec:lad}
Log file analysis enables early detection of relevant incidents such as the system failures
\cite{DBLP:conf/ijcnn/GuoYW21,DBLP:conf/icse/LeZ22,DBLP:journals/eswa/CatilloPV22,DBLP:conf/kbse/LeZ21,DBLP:journals/corr/abs-2207-03820},
aiming to identify patterns in log data and automatically notify system operators of unexpected events without manually modeled anomalies.
Specifically, 
LogBERT \cite{DBLP:conf/ijcnn/GuoYW21} 
 embed data instances into a vector space via the self-attention mechanisms over the log sequences. The hyper-sphere objective function in this model makes the distance to the center of a hyper-sphere represent the anomaly score, so that the similar instances are closer to each other than dissimilar ones. 
Le et al. \cite{DBLP:conf/icse/LeZ22} introduce a log-based anomaly detection workflow to comprehensively analyse five progressive deep learning-based models.
Note that existing methods for log data analysis typically rely on a log parser to extract the log keys (string templates) \cite{DBLP:conf/ijcnn/GuoYW21} or log events \cite{DBLP:conf/icse/LeZ22} from the original log messages, which are then used to create the ordered sequences for log 
 representation. 
However, our machine (log) data is semi-structured and multi-directional,  with a vertical direction of the time and a horizontal direction of multiple indicators extending the machine data at a single moment, which differs from the unidirectional and serial log data typically considered by these above methods.
}

\section{Conclusion}
In this paper we 
\cz{presents a novel pre-trained language model, TeleBERT, specifically designed for tele-domain to learn the general semantic knowledge.}
Meanwhile, we further introduce its improved version, KTeleBERT, which incorporates both implicit information  from the machine log data and explicit knowledge contained in our designed Tele-KG.
Experiences are summarized as causal sentence extraction, tele special token selection and prompt templates construction for unifying multi-source and multi-modal data.
Besides, we design an adaptive numeric encoder (ANEnc) for encoding fine-grained numerical data  such as the tele-indicators for KPI.
We evaluated the effectiveness of our model on three key downstream tasks in fault analysis with corresponding solutions attached: root-cause analysis, event association prediction, and fault chain tracing, demonstrating the robustness and effectiveness of our models.
 \cz{Overall, our work serves as a significant step towards improving the capability of  fault analysis  in the telecommunications industry.}


\bibliographystyle{IEEEtran}
\bibliography{main}
\clearpage
\appendix

\subsection{Details for Downstream Tasks}
\subsubsection{Problem Definition for Task1 Root-Cause Analysis}
{
The structure of the tele-system mainly refers to the connections between network elements (NEs), which is suitable to represented as a graph with NEs as nodes and their connections as edges. With the graph representation, }
the root-cause analysis aims at finding the node {in the graph} that is most likely to be the source of a fault. 
{Since the output of root-cause analysis algorithms are used to assist engineers to figure out the root-cause, similar to recommend the most likely root-cause nodes, we }
formulate this task as a node ranking problem 
{Providing a limited number and ranked list of possible root-cause nodes makes it more flexible for engineers to utilize them as clues. 
}
\subsubsection{$d_{ij}$ in Task2 Event Association Prediction}
The value of $d_{ij}$ is better with not too big or not too small, its encoding mainly aims at finding an appropriate time interval for the event relatedness. That is to say, when an event arises, and another event arises after a long time, the relatedness between them is questionable, in contrast, when another event arises after a very short time at almost simultaneously, the relationships between events may be not limited to ``trigger'' but are relations such as ``co-occurrence''. Therefore, we decided to leverage a fully connection layer by weight parameter of $W_{1}$ to automatically find the reasonable arising time interval for the current event relatedness. We are sorry to not motivate it clearly, more explanations have been added in the new manuscript. 

\subsubsection{Model Performance in Task2 Event Association Prediction}
When it comes to the reason that KTeleBERT-STL perform better than KTeleBERT-PMTL and KTeleBERT-IMTL in EAP, in our opinion, it is because the proportion of trigger rules in the entire knowledge base is not large, and part of the information in Tele-KG is unrelated and noisy to event association prediction, which could not provide informative information for event association prediction. 

\subsection{Details for Datasets Scale}
{In real tele-application, data with faults are rare. It is not possible to collect much data with faults in the same network. In the experiments, we have test our model on the dataset that are relatively large in real applications. }
{In the root-cause analysis and fault chain tracing tasks, the first baseline \textit{Random} could be regarded as an ablation experiment of KTeleBERT, where the task model randomly initialize the feature representation and train it with task data without utilizing the service vectors from pre-trained model. Compared to \textit{Random}, models utilizing vectors from KTeleBERT performs better showing the effectiveness of KTeleBERT. In Section \ref{exp:ablation}, we conduct ablation study on our proposed numerical contrastive learning method.
To further justify the advantages of ANEnc, We have involved a new task: Abnormal KPI Detection. Please refer to Sec. \ref{sec:newANE} in our latest manuscript for details.
}

\subsection{Difference between Machine Data and Typical Log Data}
\begin{figure}[!htbp]
 \centering
 \includegraphics[width = 1.0\linewidth]{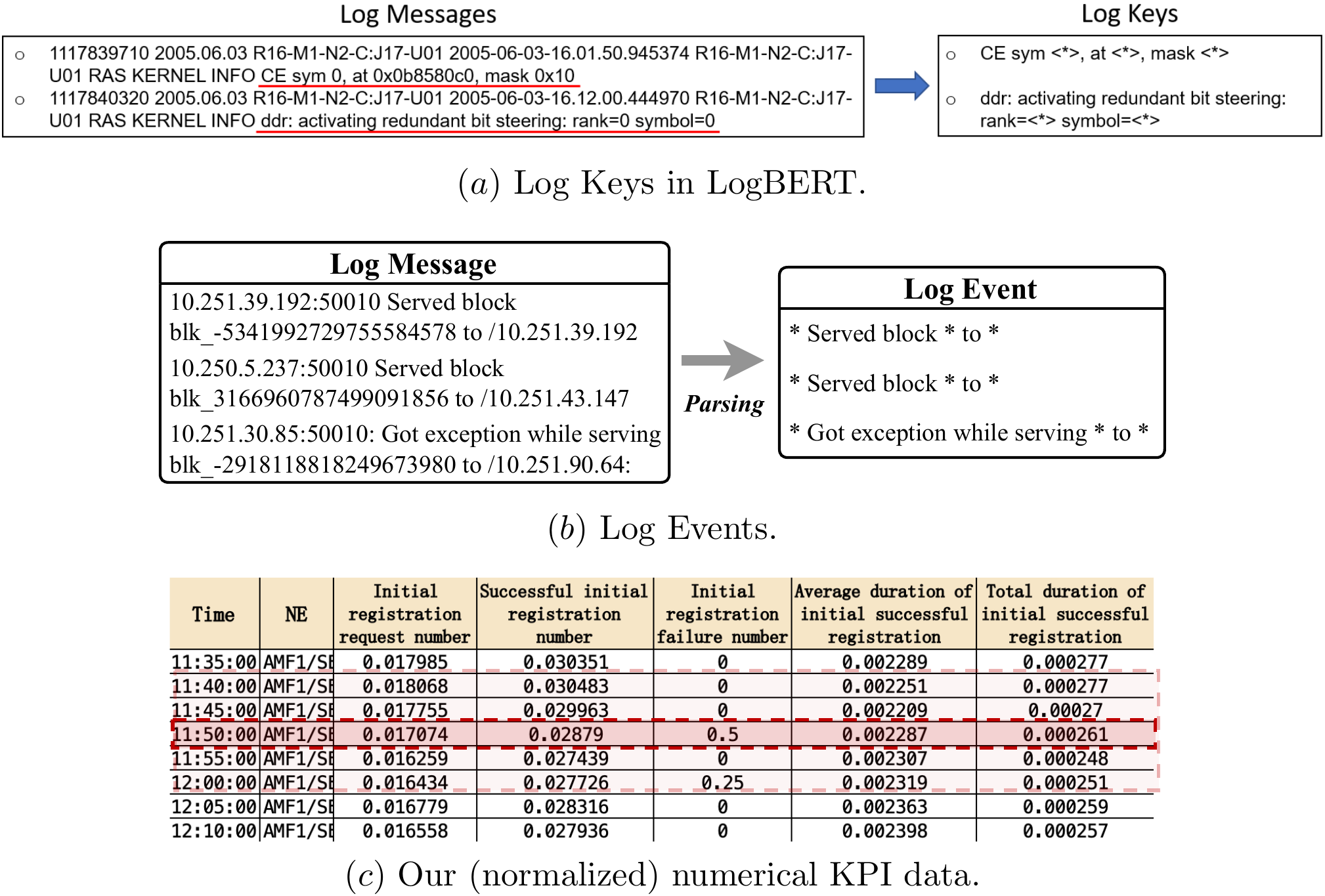}
 \caption{Log data formats compared with previous methods such as the log keys in LogBERT \cite{DBLP:conf/ijcnn/GuoYW21} and log events in \cite{DBLP:conf/icse/LeZ22}.}
 \label{fig:datadiff}
 \vspace{-5pt}
\end{figure}

We note that existing methods for log data analysis (or pre-training) typically rely on a log parser to extract the log keys (string templates) \cite{DBLP:conf/ijcnn/GuoYW21} or log events \cite{DBLP:conf/icse/LeZ22} from the original log messages. These events are then used to create an ordered sequence of logs for representation, similar to the sequential format of natural language. 
However, as shown in Fig. \ref{fig:datadiff}, our machine (log) data is semi-structured and multi-directional, with a vertical direction of the time and a horizontal direction of multiple indicators extending the machine data at a single moment. This differs from the unidirectional, serial log data typically considered by these methods.

It is also worth noting that while the alarm and KPI are import parts of the machine data in tele-domain, they are not the whole collection of machine data.
Other data sources like signaling flow and configuration data are not considered in this paper but may be studied in future work. 

\end{document}